\documentclass[conference]{IEEEtran}

\usepackage{cite} %
\usepackage{graphicx}
\usepackage{amssymb}
\usepackage{amsmath}
\usepackage{multirow}
\usepackage{comment}
\usepackage{algorithm}
\usepackage[]{algpseudocode}
\usepackage{balance}
\usepackage{mathtools}
\usepackage{bm}
\usepackage{url}
\usepackage{cleveref} %
\usepackage{xspace} %

\makeatletter
\let\MYcaption\@makecaption
\makeatother
\usepackage[font=footnotesize]{subcaption}
\makeatletter
\let\@makecaption\MYcaption
\makeatother

\DeclarePairedDelimiter\floor{\lfloor}{\rfloor}

\newcommand{\pc}{\textit{HPContext}\xspace}
\newcommand{\cfc}{\textit{HCFContext}\xspace}
\allowdisplaybreaks

\begin{document}
\title{HCFContext: Smartphone Context Inference via Sequential History-based Collaborative Filtering}

\author{
  Vidyasagar Sadhu\IEEEauthorrefmark{1}, Saman Zonouz\IEEEauthorrefmark{1}, Vincent Sritapan\IEEEauthorrefmark{2}, and Dario Pompili\IEEEauthorrefmark{1}\\
  \IEEEauthorrefmark{1}\textit{Department of Electrical and Computer Engineering, Rutgers University}, New Brunswick, USA\\
  \IEEEauthorrefmark{2}\textit{Cyber Security Division, Department of Homeland Security Science \& Technology Directorate}, USA\\
  \IEEEauthorrefmark{1}\{vidyasagar.sadhu, saman.zonouz, pompili\}@rutgers.edu, \IEEEauthorrefmark{2}vincent.sritapan@hq.dhs.gov
}

\maketitle

\thispagestyle{empty}
\pagestyle{plain} 
\pagenumbering{gobble}

\begin{abstract}
Mobile context determination is an important step for many context-aware services such as location-based services, enterprise policy enforcement, building/room occupancy detection for power/HVAC operation, etc. Especially in enterprise scenarios where policies (e.g., attending a confidential meeting only when the user is in ``Location X'') are defined based on mobile context, it is paramount to verify the accuracy of the mobile context. To this end, two stochastic models based on the theory of Hidden Markov Models~(HMMs) to obtain mobile context are proposed---\emph{personalized model} (\pc) and \emph{collaborative filtering model} (\cfc). The former predicts the current context using sequential history of the user's past context observations; the latter enhances \pc with collaborative filtering features, which enables it to predict the current context of the primary user based on the context observations of users related to the primary user, e.g., same team colleagues in company, gym friends, family members, etc.
Each of the proposed models can also be used to enhance/complement the context obtained from sensors. 
Furthermore, since privacy is a concern in collaborative filtering, a privacy-preserving method is proposed to derive \cfc model parameters based on the concepts of homomorphic encryption. Finally, these models are thoroughly validated on a real-life dataset.
\end{abstract}


\section{Introduction}\label{sec:intro}

\begin{figure}
\begin{center}
\includegraphics[width=3.3in]{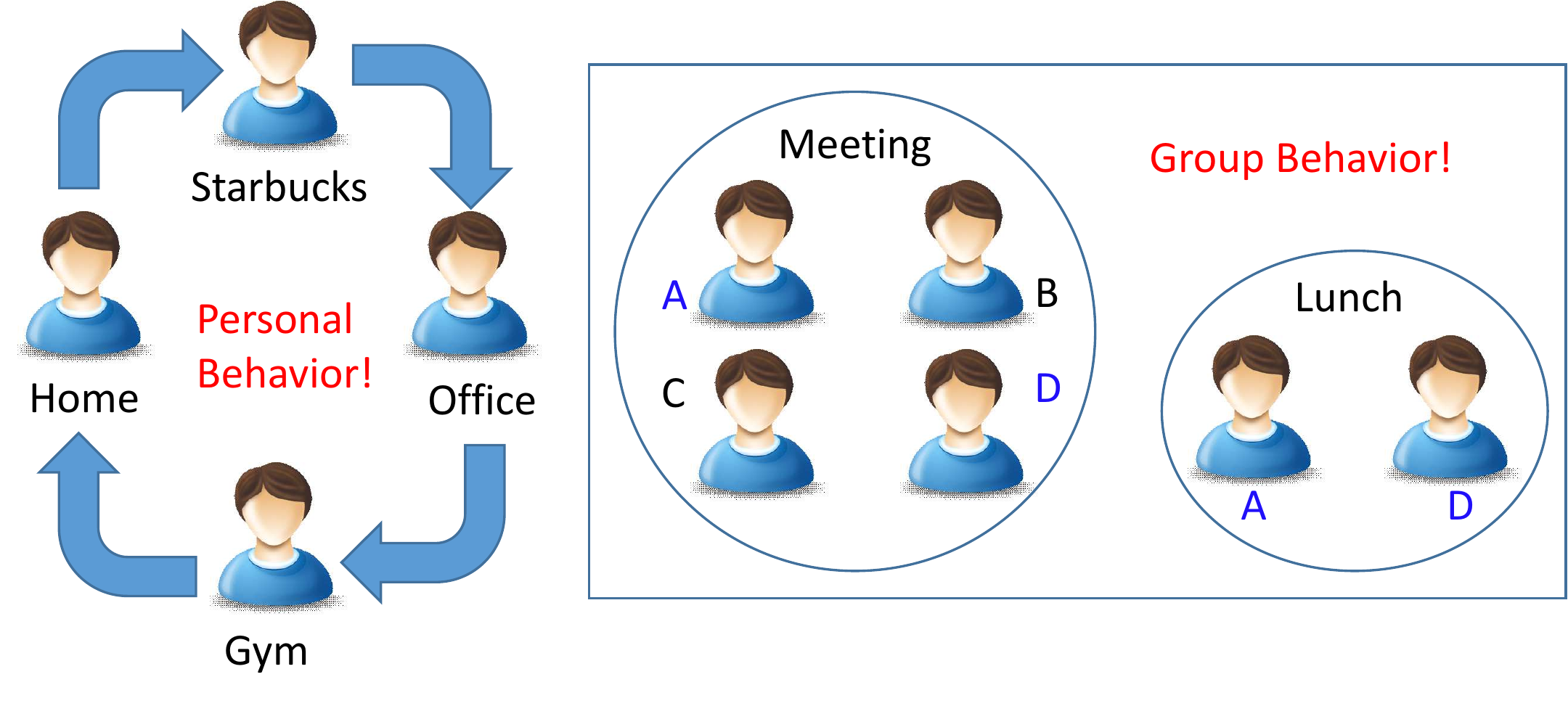}
\end{center}
\vspace{-0.1in}
\caption{A user's personal (left) and group/collaborative filtering (right) behavior that can provide some clues about user's context at any given time.}\label{fig:hcfcontext}
\vspace{-0.2in}
\end{figure}

\textbf{Overview:}
Mobile device applications provide an increasing number of features customized to match users' needs. These needs are very often inferred from specific features such as the user location, activity (e.g., running, walking, driving), surrounding people, interacting people, the current app usage on the device, etc.
These features collectively define a specific user (mobile) \textit{context}. Mobile applications are increasingly making use of these contexts such as location-based services (e.g., Foursquare, Google Now, Weather updates, etc.), enhanced reality applications (Pokemon GO~\cite{pokemon}), continuous authentication, etc. However, to enable these services, context inference is a much needed and important step (on a relevant note, see our previous work on privacy-preserving, distributed, smartphone localization framework~\cite{Sadhu2017icccn}). \textit{Most of the existing work focuses on obtaining mobile context instantaneously from sensors which could possibly be hacked, noisy or insufficient and as such cannot be relied in certain security applications. Hence we take a different approach to that problem in this paper by modeling mobile context based on past context data.}
There are many advantages of modeling the user context by leveraging the sequential nature of context information in a user's history as it can be used to predict the current or future contexts. The former can be used to validate and/or enhance the possibly hacked/noisy/insufficient sensor context, while the latter can provide some information ahead of time to the benefit of the user~\cite{anticipatorymc2015}. For example, a user's general routine during weekdays could be to head first to Starbucks near his home, then to his work and then to Gym and back to home as shown in Fig.~\ref{fig:hcfcontext}(left). \textit{The learned model will capture this behavior and can be used to validate the location of the user obtained via GPS at 5:30 pm to be at Gym (current context prediction) or display coupons related to Starbucks on his phone in advance (future context prediction).}
The latter can be leveraged by mobile personal assistant technologies such as Apple Siri/Google Now to much benefit of the user.

\textbf{Motivation:}
One of the context-aware services is the enterprise data access control as in~\cite{swirls}, where policies are defined for enterprise data access based on the phone's context (e.g., connected Wi-Fi, Cell ID, time, etc.). For example, a policy may be defined to allow the phone to be used to attend a confidential meeting or open a confidential document only when the phone's context is found to be within a given location and time. In these secure scenarios, it is not suggested to rely solely on the context obtained from the phone's sensors (e.g., GPS/WiFi and System clock to give location and time) because \textit{they could be hacked unknowingly to the user.} For example, a virus might change the system clock to show different time or spoof the GPS~\cite{Tippenhauer2011} to show different location. 
However it is hard to hack a model (more so, a collaborative one) that is learned over a long period of time. Hence our solution can be used to \textit{validate} the context directly obtained from sensors at that instant. 
Secondly, it is possible that context from sensors is \textit{noisy (due to malfunction) or does not contain enough information}. For example, in the case of a tablet or old mobile phone, it may not be able to acquire (accurate) GPS signal. As such it will be helpful if there is another way of obtaining this information such that it complements the context from sensors. 
Thirdly, as mentioned earlier, future context prediction is useful in certain context-aware services such as mobile personal assistant technologies (Google Now, Apple Siri, etc.), which can help \textit{pre-fetch information/pre-plan based on the predicted context}.

\textbf{Our Approach:}
In order to address the above issues, 
we first propose a personalized model (\pc) that predicts the user's context based on its past sequential history of contexts. Obtaining context through two approaches---sensors and personalized model---adds an extra layer of confidence to the obtained context. However, the following situations are possible: contexts obtained from both approaches are very different, contexts from one of the approaches is not available (e.g., GPS may not be available from phone sensors indoors, etc.) or insufficient leading to uncertainty. In such situations, assuming the user is closely connected to a group of people (e.g., same team colleagues in the company as shown in Fig.~\ref{fig:hcfcontext}(right), gym friends, family members, etc.),
\textit{it is possible that the context of other members in that group of people can provide additional information about his/her context.} 
For example, assume users $A$ and $D$ often go to lunch together (learned via model). Now somehow if it is known that $D$ is going to ``Restaurant1'' tomorrow for lunch, it is most likely that the context of user $A$ tomorrow around 1 pm is ``having lunch with $D$ at Restaurant1" without having to rely on $A$'s phone sensors at that instant. Our paper explores this aspect of context to provide a second layer of confidence to the context (over and above the personalized model).
For this purpose, we propose to use such context obtained through \textit{collaborative filtering} of the contexts of users closely related to the primary user (\cfc). To the authors' best knowledge, this is the first work to explore collaborative filtering for mobile contexts that can be used to \textit{validate} and/or \textit{enhance} the \textit{current} context obtained from sensors or predict the \textit{future} context for mobile personal assistant technologies.
Additionally we present a privacy-preserving method for parameter estimation (training) of \cfc, as users may not be willing to share their private data with each other for the same. 

\textbf{Contributions:}
Our specific contributions are as follows.
\begin{itemize}
\item We propose a personalized (\pc) and collaborative filtering (\cfc) model to \textit{predict} the users context at any given instant (including future) based on the sequential history of past contexts and based on the theory of Hidden Markov Models~(HMMs). We design a novel emission model for these HMMs by considering the unique features and the nitty-gritties of a mobile context (e.g., GPS from sensors may not be always available). 
\item We present a homomorphic encryption based privacy-preserving approach for training the \cfc.
\item We validate the efficacy of the proposed models by testing them on a real-life data set belonging to five graduate students collected over two months. We also evaluate our privacy-preserving approach to study its trade-offs.
\end{itemize}

\textbf{Paper Outline:}
In Sect.~\ref{sec:rel-work}, we present the related work and position our paper. In Sect.~\ref{sec:prop-soln}, 
we discuss the proposed models, (\pc, \cfc) and the privacy-preserving approach for the parameter estimation of \cfc.
In Sect.~\ref{sec:eval}, we present the results of our proposed approaches. 
Finally, in Sect.~\ref{sec:conc}, we conclude and discuss future directions.

\section{Related Work}\label{sec:rel-work}
In this section, we position our work with respect to previous works (i) that obtain context with and without users sequential history, (ii) via local collaborative sensing, and (iii) related to privacy-preserving collaborative filtering.

\textbf{Without Sequential History:}
There is existing work on modeling mobile contexts without considering the sequential nature of context information. For example, Bao et al.~\cite{Bao2010} propose an unsupervised approach to model mobile context from raw contextual data using Latent Dirichlet Allocation~(LDA). Srinivasan et al.~\cite{Srinivasan2014} mine the co-occurrences of certain context attributes; frequently and simultaneously occurring context attributes are formulated as association rules to predict what else the user will do (e.g., read comics) given a current context attribute (e.g., listen to jazz). However, unlike ours these approaches do not exploit temporal dependencies among 
contexts but only consider the behavior at a given instant. 

\textbf{With Sequential History:}
There is also work that exploits the sequential/temporal dependencies between contexts. For example, Mukherji et al.~\cite{Mukherji2014} present Mobile Sequence Miner~(MSM) framework that mines frequent sequences occurring in app usage patterns, location visits, and call logs using a frequency-based approach. 
Farrahi et al.~\cite{Farrahi2014} present a probabilistic approach to mine mobile phone data (e.g., location) sequences using Distant N-Gram Topic Model~(DNTM) where they model the sequence to be dependent on the starting element of the sequence. There are works that model the user activity using HMM based on sensor measurements~\cite{Mannini2010,JonathanFeng-shunLin2011}. 
Even though these approaches exploit the sequential nature of contextual information, they neither consider collaborative filtering nor privacy-preserving aspects like we do. We claim that collaborative filtering context has additional context information 
than context obtained from personal history alone.

\textbf{Local Collaborative Sensing:}
There are also works on local collaborative sensing; however, these works do not consider the sequential nature of past information into the collaboration process~\cite{audiocollab2014}. 
For example, 
Mantyjarvi et al.~\cite{collabcxtrecog2003} present a collaborative sensing approach where a device, upon noticing a change in its local context beyond a threshold value, requests contexts from its surrounding devices so as to increase the accuracy of its context vector. Miluzzo et al.~\cite{darwinphones2010} use collaboration to increase the confidence of the sensed context through consensus of contexts sensed at surrounding devices. These approaches however do not consider past sequential nature of context information into collaboration.

\textbf{Privacy-preserving Collaborative Filtering:}
There is existing literature in the domain of privacy-preserving collaborative filtering and HMM techniques, which can be broadly classified into two categories---\emph{data perturbation/randomization} to hide the original data albeit with accuracy loss and \emph{data encryption} with typically no accuracy loss albeit with higher computational complexity. On the former, Polat et al.~\cite{Polat2003} and Parameswaran et al.~\cite{Parameswaran2007} present privacy-preserving collaborative filtering techniques based on randomized perturbation and data obfuscation respectively.
On the latter, Guo et al.~\cite{Guo2007} present a privacy-preserving Markov model for sequence classification using homomorphic and ElGamal cryptographic systems. 
More works in this category can be found in~\cite{Renckes2007,Kikuchi2009,Nguyen2013}.
We present an approach, designed specifically for our scenario, that extends the ideas in this category for privacy preserving multi-party parameter estimation of our \cfc model.

\section{Proposed Approach}\label{sec:prop-soln}
In this section, we describe \pc and \cfc models  (Sect.~\ref{sec:prop-soln:collab_filt}) and
a privacy-preserving approach for parameter estimation of \cfc (Sect.~\ref{sec:prop-soln:privacy}).

\begin{table}[t]
\footnotesize
\caption{Example Context Observations of User $u$ for Times $t=1,...,T$.}\label{table:context_obs}
\vspace{-0.15in}
\begin{center}
\begin{tabular}{|p{1cm}|p{7cm}|} 
 \hline
 \textbf{Time ($t$)} & \textbf{Context observation of user $u$ at time $t$ ($\bm{O}_{tu}$)} \\ 
 \hline
 \hline
 \multirow{2}{4em}{$t_1$} & WiFi: \textit{wifi1}, CellID: \textit{cid1}, LAC: \textit{lac1}, Battery Level: \textit{high}, Battery Status: \textit{discharging}, Day Period: \textit{morning}, Day of week: \textit{Monday}, Holiday: \textit{No}  \\ 
 \hline
 \multirow{2}{4em}{$t_2$} & WiFi: \textit{wifi2}, CellID: \textit{cid2}, LAC: \textit{lac2}, Battery Level: \textit{medium}, Battery Status: \textit{discharging}, Day Period: \textit{noon}, Day of week: \textit{Monday}, Holiday: \textit{No}  \\ 
 \hline
 \multicolumn{2}{|c|}{...........} \\
 \hline
 \multirow{2}{4em}{$t_T$} & WiFi: \textit{wifi1}, CellID: \textit{cid1}, LAC: \textit{lac1}, Battery Level: \textit{low}, Battery Status: \textit{charging}, Day Period: \textit{night}, Day of week: \textit{Sunday}, Holiday: \textit{Yes}  \\ 
\hline
\end{tabular}
\vspace{-0.3in}
\end{center}
\end{table}

\subsection{Context from Collaborative Filtering (\cfc)}\label{sec:prop-soln:collab_filt}
\begin{figure}
\begin{center}
\includegraphics[width=3.1in]{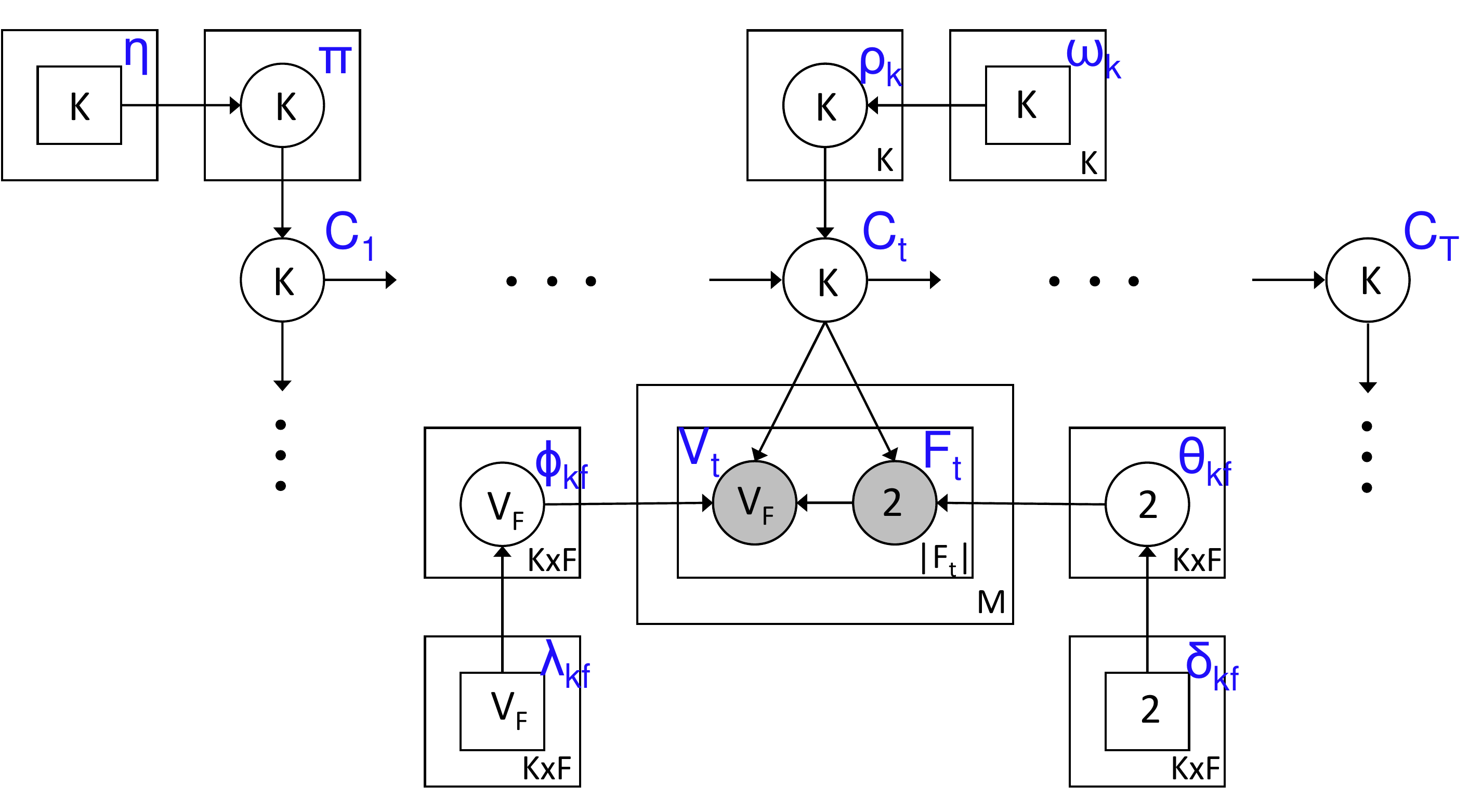}
\end{center}
\vspace{-0.1in}
\caption{Plate notation of \cfc. $F$ is the total number of features, $V_F$ denotes the total number of possible values for feature $F$, $|F_t| = |V_t|$ denotes the number of features observed at time $t$, $M$ is the number of users, and $K$ is the number of hidden context states.}\label{fig:mohmm}
\vspace{-0.2in}
\end{figure}

\noindent \textbf{Problem Formulation}: We present here the proposed \cfc model \emph{by designing a novel emission model of a HMM taking into account the multi-user collaborative filtering aspects, as well as the unique features of the mobile context and its nitty-gritties such as feature unavailability}. 
We first start with notation---capital letters denote random variables, whereas their small equivalents are their realizations. A vector variable will be indicated in bold. We model context ($C_t$) as a latent variable in the HMM. For a given user, the observation corresponding to a context state at time $t$ will be called context observations ($\bm{O}_t$). An example of a user's context observations from time $t = 1...T$ is shown in Table~\ref{table:context_obs}. Each observation, $\bm{O}_t$, consists of a set of contextual feature-value pairs. These observations are obtained at regular time intervals (e.g., a minute to four hours). It can be seen as an example from the table that the context observation at $t=t_1$ corresponds to morning when the user is at home (battery is high, probably because the user charges her phone the previous night). The observation at $t=t_2$, say after 4 hours, can be interpreted as being at office or workplace (change of WiFi, Cell ID, etc.) with battery level being in medium range. Finally, the observation at $t=t_T$ (after several days) can be taken to be again at home in the night. We assume that $K$ number of latent context states spans across these $T$ observations. Considering users $u = 1,...,M$, each user has a similar set of $T$ observations. Plate notation~\cite{platenotation} for our \cfc model for $M$ users is shown in Fig.~\ref{fig:mohmm}. In plate notation, the number of different categorical values a random variable can take is shown inside the circle or rectangle. A circle is used for a random variable while a rectangle is generally used for hyperparameters. Observable variables are shaded. The number of repetitions of a rectangular block is shown at its bottom right corner. 
For a given user $u$, the observation at time $t$, $\bm{O}_{tu}$ is a set of feature-value pairs (as in a row of Table~\ref{table:context_obs}). We can write $\bm{O}_{tu}=(\bm{f}_{tu},\bm{v}_{tu}) = (f_{t,u,i},v_{t,u,i})_{i=1}^{|f_{tu}|}$, where $|f_{tu}|$ is the number of available features of user $u$, at time $t$. Generation of each variable in Fig.~\ref{fig:mohmm} is described next.

\noindent\underline{\textit{Initial State Model:}} A prior distribution of \textit{contexts}, $\pi$ is generated from prior Dirichlet distribution, $\eta$. $C_1$ is then generated from $\pi$. We will assume a total of $K$ possible context states for \cfc over all $M$ users.\\
\underline{\textit{Transition Model:}} A prior \textit{transition} distribution of contexts, $\rho_{c_{t-1}}=\rho_k$,
is generated from a prior Dirichlet distribution, $\omega_k$. $C_t$ is then generated from $\rho_{c_{t-1}}$ for a given $C_{t-1}$. Note that $\rho_k$ and $\omega_k$ can take a total of $K$ categorical values ($C_t$, current state) for each $k$ ($C_{t-1}$, previous state).\\
\underline{\textit{Novel Emission Model:}} For $\bm{O}_t$ generation under each $C_t$, since features are not always available (e.g., GPS is not available when indoor or underground, etc.), we will define a separate distribution for features ($F_t$) to account for their availability and then another distribution to obtain the values ($V_t$) for those features at time $t$, as illustrated in Fig.~\ref{fig:mohmm}. $F_t$ is dependent on $C_t$, whereas $V_t$ is dependent on both $C_t$ and $F_t$. An initial \textit{feature} distribution, $\theta_{c_t,f_t} = \theta_{k,f}$, is generated from a prior Dirichlet distribution, $\delta_{k,f}$. Feature $F_t$ is then generated from $\theta_{k,f}$, which can take two categorical values---whether the feature is present or not for the given context, $c_k$. We will assume a total of $F$ possible features over all observations of all users. A prior \textit{value} distribution, $\phi_{c_t,f_{t}} = \phi_{k,f}$, is generated from a prior Dirichlet distribution, $\lambda_{k,f}$. Value $V_t$ is then generated from $\phi_{c_t,f_{t}}$ for a given context $c_t$ and feature $f_{t,i}$. For a given feature $f$, we will assume $V$ can take $V_f$ possible values.
The priors will be chosen such that the summation and non-negativity constraints on the parameters are satisfied and also to encode prior information. For example, if it is known that a user frequently moves between home to work, the prior parameter for this transition, ($\omega_k$), is given a high value. 

\noindent \textbf{Parameter Estimation (Training):}
Given these parameters, $\bm{\Psi} = \{\Pi, P, \Theta, \Phi \}$, the parameter space of $\pi, \rho_k, \theta_{k}, \phi_{k,f}$ and their hyperparameters, $\{\eta, \omega, \delta, \lambda \}$, let us denote the joint probability of all latent contexts $\bm{C} = \{ C_1,...C_T \}$ and all context observations, $\bm{O} = \{ O_1,...O_T \}$, as $P(\bm{C,O} \mid \bm{\Psi},\eta, \omega, \delta, \lambda)$.
\begin{comment}
as follows,
\begin{align}\label{eq:joint}
&P(\bm{C,O} \mid \bm{\Psi},\eta, \omega, \delta, \lambda) = \int {P(\pi \mid \eta)} {P(c_1 \mid \pi)} d\pi \nonumber \\ 
&\times \int \prod_{k=1}^K {P(\rho_k \mid \omega_k)} \prod_{t=2}^T {P(c_t \mid c_{t-1},\rho_{c_{t-1}})} d\rho \nonumber \\
&\times \int \prod_{k=1}^K \prod_{f=1}^F {P(\theta_{k,f} \mid \delta_{k,f})} \prod_{u=1}^M \prod_{t=1}^T \prod_{i=1}^{2}  {P(f_{t,i,u} \mid c_t, \theta_{c_t})}  d\theta \nonumber \\
&\times \int \prod_{k=1}^K \prod_{f=1}^F {P(\phi_{k,f} \mid \lambda_{k,f})} \nonumber \\
&\prod_{u=1}^M \prod_{t=1}^T \prod_{i=1}^{V_f} {P(v_{t,i,u} \mid c_t, f_{t,i,u}, \phi_{c_t, f_t})}  d\phi. 
\end{align}
\end{comment}
Likelihood of the observations $\bm{O}$ is then,
\begin{equation}\label{eq:likelihood}
L(\bm{O}) = \sum_{\bm{c}} {P(\bm{C,O} \mid \bm{\Psi},\eta, \omega, \delta, \lambda)}.
\end{equation}
Training the HMM involves finding the parameters $\bm{\Psi}$ that maximize the likelihood in~\eqref{eq:likelihood}. Given the complex nature of~\eqref{eq:likelihood} expanded, it is very difficult to derive a closed-form solution of $\bm{\Psi}$. Hence, we make use of the well-known iterative approach called Expectation Maximization~(EM)~\cite{Bishop2006} \textit{but modify it to fit our approach}. It consists of two important relations viz., forward and backward relations.
We can express the full observation probability $p(\bm{o}_t \mid c_{tk}) = \mu_{tk}$:
\begin{equation}
\mu_{tk} = p(\bm{o}_t \mid c_{tk}) = \prod_{u=1}^M \mu_{tk_u}= \prod_{u=1}^M \prod_{\substack{f \in \bm{f}_{tu} \\ v \in \bm{v}_{tu}}} \theta_{k,f}~\phi_{k,f,v}.  \label{eq:mu}
\end{equation}
Forward relation can now be expressed as,
\begin{align}
\alpha(c_{1k}) &= \pi_k \mu_{1k}~\textit{(for t = 1)}, \label{eq:alpha2} \\
\alpha(c_{tk}) &= \mu_{tk} \sum_{j=1}^K \alpha(c_{t-1,j}) \rho_{jk}~\textit{(for t = 2 to T)}. \label{eq:alpha3}
\end{align}
Similarly, we can express the backward relation as follows,
\begin{equation}\label{eq:beta2}
\beta(c_{tk}) = \sum_{j=1}^K \beta(c_{t+1,j}) \mu_{t+1,j} \rho_{kj}.
\end{equation}

\noindent Using these relations, we can define two new variables, $\xi(C_{t-1},C_t)$ and $\gamma(C_t)$, for ease of analysis (denote $\gamma(C_t = c_{tk}) =\gamma(c_{tk})$ and similarly for $\xi(c_{t-1,j},c_{tk})$) as follows,
\begin{align}
\xi(c_{t-1,j},_{tk}) &= \frac{\alpha(c_{t-1,j}) \mu_{tk} \rho_{jk} \beta(c_{tk})}{\sum_{k=1}^K \alpha(c_{Tk})}, \label{eq:xiparams}\\
\gamma(c_{tk}) &= \sum_{j=1}^K \xi(c_{t-1,j},c_{tk}). \label{eq:gammaparams}
\end{align}

\noindent We can now compute the model parameters as,
\noindent\begin{flalign}
\hspace{-0.2in}
\pi_k &= \frac{\gamma(c_{1k}) + \eta_k}{\sum_{k'=1}^K (\gamma(c_{1k'}) + \eta_k')}, \label{eq:param1}\\
\hspace{-0.2in}
\rho_{kj} &= \frac{\sum_{t=2}^T \xi(c_{t-1,k},c_{tj}) + \omega_{kj}}{\sum_{j'=1}^K \sum_{t=2}^T \xi(c_{t-1,k},c_{tj'}) + \sum_{j'=1}^K \omega_{kj'}}, \label{eq:param2}\\
\hspace{-0.2in}
\theta_{k,f} &= \frac{\sum_{t=1}^T \gamma(c_{tk}) \sum_{u=1}^M \bm{I}(f \in \bm{f}_{tu}) + \delta_{k,f}}{M\sum_{t=1}^T \gamma(c_{tk}) + \sum_{f'=1}^F \delta_{k,f'}}, \label{eq:param3}\\
\hspace{-0.2in}
\phi_{k,f,v} &= \frac{\sum_{u=1}^M \sum_{t: f \in \bm{f}_{tu}} \gamma(c_{tk}) \bm{I}(v_{t,f,u} = v) + \lambda_{k,f,v}}{\sum_{u=1}^M \sum_{t: f \in \bm{f}_{tu}} \gamma(c_{tk}) + \sum_{v'=1}^{V_f} \lambda_{k,f,v'}}, \label{eq:param4}
\end{flalign}where $\bm{I}(x)$ is the indicator function with $\bm{I}(x) = 1$ if $x$ is true and $0$ otherwise. Eqs.~\eqref{eq:alpha2}-\eqref{eq:gammaparams} constitute the E-step, while Eqs.~\eqref{eq:param1}-\eqref{eq:param4} constitute the M-step of the EM algorithm. These steps are iterated until the parameters in M-step converge.

\noindent \textbf{Prediction:}
We will now use the learned parameters to predict the future observations given past observations. This will be done by first finding the distribution over future states and then by multiplying the distribution over the observations given the future state. In our case, since the observations are feature-value pairs, we will first calculate the distribution over features and then the distribution over values given features. Given that the user has made a sequence of past `t' observations, $\bm{o}_{1:t} = \{\bm{o}_1,...,\bm{o}_t\}$, the probability that a feature $f$, and then a value $v$ for that feature, will be observed at time $t+1$ can be computed, respectively, as follows,
\begin{align}
p(f \in \bm{f}_{t+1} \mid \bm{o}_{1:t}) &= \sum_{k=1}^K p(c_{t+1,k} \mid \bm{o}_{1:t}) \cdot \theta_{k,f}, \label{eq:featurepred}\\
p(v_{t+1,f} = v \mid \bm{o}_{1:t}) &= \sum_{k=1}^K p(c_{t+1,k} \mid \bm{o}_{1:t}) \cdot \phi_{k,f,v}, \label{eq:valuepred}
\end{align}

Note that $p(c_{t+1,k} \mid \bm{o}_{1:t})$ in~\eqref{eq:featurepred}, \eqref{eq:valuepred} is computed as,
\begin{equation}
p(c_{tk} \mid \bm{o}_{1:t}) = \sum_{j=1}^K p(c_{t+1,k} \mid c_{tj}) p(c_{tj} \mid \bm{o}_{1:t}),
\end{equation}
where $p(c_{t+1,k} \mid c_{tj}) = \rho_{jk}$ and $p(c_{tj} \mid \bm{o}_{1:t})$ can be recursively computed using a procedure similar to~\eqref{eq:alpha3},
\begin{equation*}
\alpha'(c_{tk}) = p(c_{tk} \mid \bm{o}_{1:t}) 
= \frac{\mu_{tk} \sum_{j=1}^K \rho_{jk} \alpha'(c_{t-1,j})}{\sum_{k=1}^K \mu_{tk} \sum_{j=1}^K \rho_{jk} \alpha'(c_{t-1,j})} \label{eq:alphaprime}
\end{equation*}

\noindent We compute~\eqref{eq:featurepred}, \eqref{eq:valuepred} over all features, $f_i : i = \{1,...,F\}$, all corresponding values $v_j : j = \{1,...,V_{f_i}\}$ and pick the most probable ones to get the feature-value pairs at $t+1$.

\noindent \textbf{Determining the Number of Hidden Contexts:}
So far we have assumed that the number of hidden contexts $K$ is given; however, in general this number needs to be determined automatically from the data. We will now detail an original approach to determine the best $K$ assuming it lies in the range $\mathcal{K}_{range}=[K_{min},K_{max}]$ and that the extremes can be approximately obtained from prior information about the data. To determine the best $K \in \mathcal{K}_{range}$, we will define a metric called \textit{Perplexity} that determines how well the chosen $K$ fits for prediction tasks over the testing set.
Finding perplexity involves calculating the prediction probability over a sequence of observations given a sequence of past observations from the testing set. Perplexity can be defined as follows,
\begin{equation}
\text{Perplexity} = \exp\bigg(-\frac{\log p(\bm{o}_{t+1:T} \mid \bm{o}_{1:t})}{\sum_{u=1}^M \sum_{t=t+1}^T~|\bm{o}_{tu}|}\bigg),
\end{equation}
where $|\bm{o}_{tu}|$ is the number of features observed at time $t$ for user $u$ and $p(\bm{o}_{t+1:T} \mid \bm{o}_{1:t})$ can be determined as follows,
\begin{equation}
p(\bm{o}_{t+1:T} \mid \bm{o}_{1:t}) = \sum_{k=1}^K p(c_{tk} \mid \bm{o}_{1:t}) p(\bm{o}_{t+1:T} \mid c_{tk}),
\end{equation}where $p(c_{tk} \mid \bm{o}_{1:t}) =\alpha'(c_{tk})$ 
and $p(\bm{o}_{t+1:T} \mid c_{tk})=\beta(c_{tk})$.
Intuitively, a small perplexity is desired. Although in general the perplexity reduces as $K$ increases, a large $K$ is not preferred due to the risk of overfitting. Hence, we make use of the rate of decrease in perplexity to determine when to stop increasing $K$, e.g., if it falls below a threshold (say 10\%).

\noindent \textbf{\pc Model:}
Personalized model is just a special case of the above collaborative filtering model when the number of users is one (i.e., $M=1$). The main difference is in the observation probability. Specifically, for user $u$, 
\begin{equation}
\mu_{tk} = p(\bm{o}_{tu} \mid c_{tk}) = \mu_{tk_u}= \prod_{\substack{f \in \bm{f}_{tu} \\ v \in \bm{v}_{tu}}} \theta_{k,f}~\phi_{k,f,v}.  \label{eq:mu2}
\end{equation}
Remaining equations remain the same with setting $M=1$.

\subsection{Privacy-preserving Multi-party Computing} \label{sec:prop-soln:privacy}
\textit{Since the users could possibly be in different contexts while training it is important to preserve their privacy.} In this section, we develop algorithms for multi-party parameter estimation of \cfc (Sect.~\ref{sec:prop-soln:collab_filt}) while preserving each party's privacy. Hence model parameters need to be jointly estimated when the individual observations are encrypted. Since the model parameters $\bm{\Psi}$ are known to everyone,
the prediction can be carried out by each party on their own. The algorithms we developed for this (extended from~\cite{Nguyen2013}) are based on the following well-known primitives---\textit{homomorphic encryption}~\cite{Paillier1999}, \textit{secure logsum}, and \textit{secure negation}~\cite{Pathak2011}.
\begin{algorithm}[t!]
\footnotesize
\caption{Secure Multi-party Computation of $P(\bm{O \mid \Psi})$.}\label{alg:forward}
\begin{algorithmic}[1]
\Require Parties $P_1, P_2, \ldots, P_M$ know the model $\bm{\Psi}$. Each party has a set of private observations $\bm{O}_{tu}$=$(\bm{f}_{1u},\bm{v}_{1u}), (\bm{f}_{2u},\bm{v}_{2u}),\ldots, (\bm{f}_{Tu},\bm{v}_{Tu})$.
\Ensure $P(\bm{O} \mid \bm{\Psi},\eta, \omega, \delta, \lambda) = \sum_{k=1}^K \alpha(c_{Tk})$

 \textbf{\underline{Initialization}~(t=1):}
  \For{$k = {1,\ldots,K}$}
    \ForAll{$P_{q\neq m}$}
      \State $P_q$ sends $E[\log(\mu_{1k_x})]$ to $P_m$,
    \EndFor
    \State $P_m$ computes $E[\log(\prod_{u=1}^M \mu_{1k_u})]$ using \eqref{eq:mu} \par
    \State $P_m$ computes $E[\log(\alpha(c_{1k}))]$ using \eqref{eq:alpha2} \par
  \EndFor

\textbf{\underline{Induction}:}
  \For{$t = {2,\ldots,T}$}
    \State Repeat steps 2-6, replacing time index with $t$, so that $P_m$ obtains $E[\log(\prod_{u=1}^M \mu_{tj_u})]$ for $j = 1,\ldots, K$.
    \For{$k = {1,\ldots,K}$}
      \State $P_m$ and $P_1$ use the secure logsum protocol to compute $E[\log \sum_{j=1}^{K}\alpha(c_{t-1,j})\rho_{jk}]$,
      \State $P_m$ computes $E[\text{log}(\alpha(c_{tk}))]$ using \eqref{eq:alpha3} \par
    \EndFor
  \EndFor
  
 \textbf{\underline{Termination}:}
  \State $P_m$ and $P_1$ use the secure logsum protocol to compute $E[\log P(\bm{O \mid \Psi})] = E[\log \sum_{k=1}^K\alpha(c_{Tk})]$,
  \State $P_1$ decrypts the result and sends the value to $P_m$.
\end{algorithmic}
\end{algorithm}
\setlength{\textfloatsep}{0pt}%

The proposed algorithm for secure multi-party computation of data likelihood, $P(\bm{O \mid \Psi})$ is shown in self-explanatory Algorithm~\ref{alg:forward}. It details the steps the $M$ parties need to undertake to jointly compute $\alpha(\cdot)$ (as per~\eqref{eq:alpha2}, \eqref{eq:alpha3}) and the log likelihood of their observation data given the model parameters, $P(\bm{O \mid \Psi})$.
We assume only one party ($P_1$ in Algo.~\ref{alg:forward}) has both the private and public keys, while the remaining $M-1$ parties ($P_m \mid m = 2,\ldots,M$) have only the public key. Hence all parties encrypt their private observation data and send it to $P_m$, where $m$ can be any one of $2,\ldots,M$, which does the computations on this private encrypted data (as it cannot decrypt it since it has only public key). Whenever it needs to compute secure logsum and secure negation protocols, it consults $P_1$ which has the private key. Algo.~\ref{alg:forward} can be similarly used to compute $\beta(\cdot)$ as per~\eqref{eq:beta2}.  The proposed algorithm for secure multi-party estimation of model parameters, $\bm{\Psi}$ is shown in Algo.~\ref{alg:bw}. It details the steps taken by the $M$ parties to estimate $\bm{\Psi}$ using $P(\bm{O \mid \Psi}), \alpha(\cdot), \beta(\cdot)$ computed from Algo.~\ref{alg:forward}. In Algo.~\ref{alg:bw}, $P_m$ first computes $\xi(c_{t-1,j},c_{tk}), \gamma(c_{tk})$ as per \eqref{eq:xiparams} and \eqref{eq:gammaparams} respectively. It then computes the model parameters, $\bm{\Psi}$ as per~\cref{eq:param1,eq:param2,eq:param3,eq:param4}. The computational complexity of Algo.~\ref{alg:forward} can be seen as $O(MK^2T)$ due to twice-nested for-loop operating for T timesteps for each party, while it is $O(MK^3T)$ for Algo.~\ref{alg:bw} due to triple-nested for-loop. 

\begin{algorithm}[t!]
\footnotesize
 \caption{Secure Multi-party Estimation of $\bm{\Psi}$.}\label{alg:bw}
\begin{algorithmic}[1]
\Require $E[\log \alpha(c_t)]$, $E[\log \beta(c_t)]$, and $E[\log P(\bm{O \mid \Psi})]$. 
\Ensure The updated model parameters $\bm{\Psi} = \{\Pi, P, \Theta, \Phi \}$
\For{$t = 2,\ldots,T$}
  \For{$k = {1,\ldots,K}$}
    \For{$j = {1,\ldots,K}$}
    \State $P_m$ computes $E[\log\xi(c_{t-1,j},c_{tk})]$ by taking the $\log$ of~\eqref{eq:xiparams} \par
    \EndFor
  \State $P_m$ and $P_1$ use the secure logsum to compute $E[\log \gamma(c_{tk})]$ from~\eqref{eq:gammaparams} \par
  \EndFor
\EndFor
\State $P_m$ uses~\eqref{eq:param1}, \eqref{eq:param2} to update $E[\log {\pi}_k], E[\log \rho_{kj}]$. \par

\State $P_m$ then updates $E[\log \theta_{k,f}]$ and $E[\log \phi_{k,f,v}]$ as in~\eqref{eq:param3} and \eqref{eq:param4}, for the $M$ parties
using the secure logsum and negation protocols.
\end{algorithmic}
\end{algorithm}
\setlength{\textfloatsep}{3pt}%

\noindent \textbf{Threat/Adversary Model:}
We use a semi-honest setting, where parties keep all their intermediate computations private, and we assume that $P_m$ will not collude with $P_1$ and disclose encrypted values received. The key generation in our security model will be following a standard key exchange mechanism~\cite{chopra2015} without the need to a third party entity. 
One may argue that the models themselves may be unreliable as they are based on historical sensor data. Our approach provides protection against this point by comparing sensor context with that obtained from \pc and \cfc, and alerting the user in case of significant differences and adaptively learning to ignore false positives.

\noindent \textbf{Floating Point and Negative Numbers:}
Our algorithms need to encrypt the HMM parameters, which are real numbers. We translate between floating-point numbers and non-negative integers by scaling and rounding off the values. Let $c$ be the scaling factor. A real number $r$ is translated to integer $\bar{r} = \floor*{cr}$, where $\floor*{x}$ is the largest integer $\leq x$. We incorporate this operation into the encryption and decryption as $E'[r] = E[\bar{r}] = E[\floor*{cr}],~and~D'[E'[r]] = \bar{r}/c \approx r$. 
For negative numbers, we use modulo $n$ arithmetic, i.e., negative numbers are represented by their modular additive inverse. For $r<0$, $E'[\bar{r}] = E'[\bar{r}+n]$. This means our $r$ is limited to range $[-n/(2c), (n-1)/(2c)]$.

\begin{table}[t]
\vspace{-0.05in}
\footnotesize
\caption{Lifemap Dataset Analysis (9 Weeks, 5 Users, 1 Hr Sampling).} \label{table:lifemap_features}
\vspace{-0.15in}
\begin{center}
\begin{tabular}{|p{1.7cm}|p{5cm}|p{0.3cm}|}
\hline
\textbf{Feature}    & \textbf{Values}    & \textbf{$V_F$}   \\ \hline \hline
WiFi                        & MAC values of $\max~\rm{dBm}$ Access Points                                                            & 440                \\ \hline
Place Name                  & User defined place name values                                                                     & 168                \\ \hline
Cell ID                     & Cell ID values                                                                                     & 316                \\ \hline
LAC                         & Location Area Code values                                                                          & 33                 \\ \hline
Batt. Level               & Low ($<35\%$), Medium ($35\% - 65\%$), High ($65\% - 85\%$), Full ($>85\%$)                        & 4                  \\ \hline
Batt. Status              & Charging, Discharging, Full                                                                        & 4                  \\ \hline
Day Period                & Morning: \{7 to 11~am\}; Noon: \{11~am to 2~pm\}; Afternoon: \{2 to 6~pm\}; Evening: \{6 to 9~pm\}; Night: \{9~pm to 7~am\} & 5 \\ \hline
Day Name                 & Mon, Tue, Wed, Thu, Fri, Sat, Sun                                                                  & 7                  \\ \hline
Holiday                     & Yes, No                                                                                            & 2                  \\ \hline
\end{tabular}
\vspace{-0.05in}
\end{center}
\end{table}

\begin{figure*}[t!]
        \centering
        \hspace{-0.3in}
           \begin{subfigure}[b]{0.32\textwidth}
        		\centering
        		\includegraphics[width=1.15\textwidth,height=1.8in]{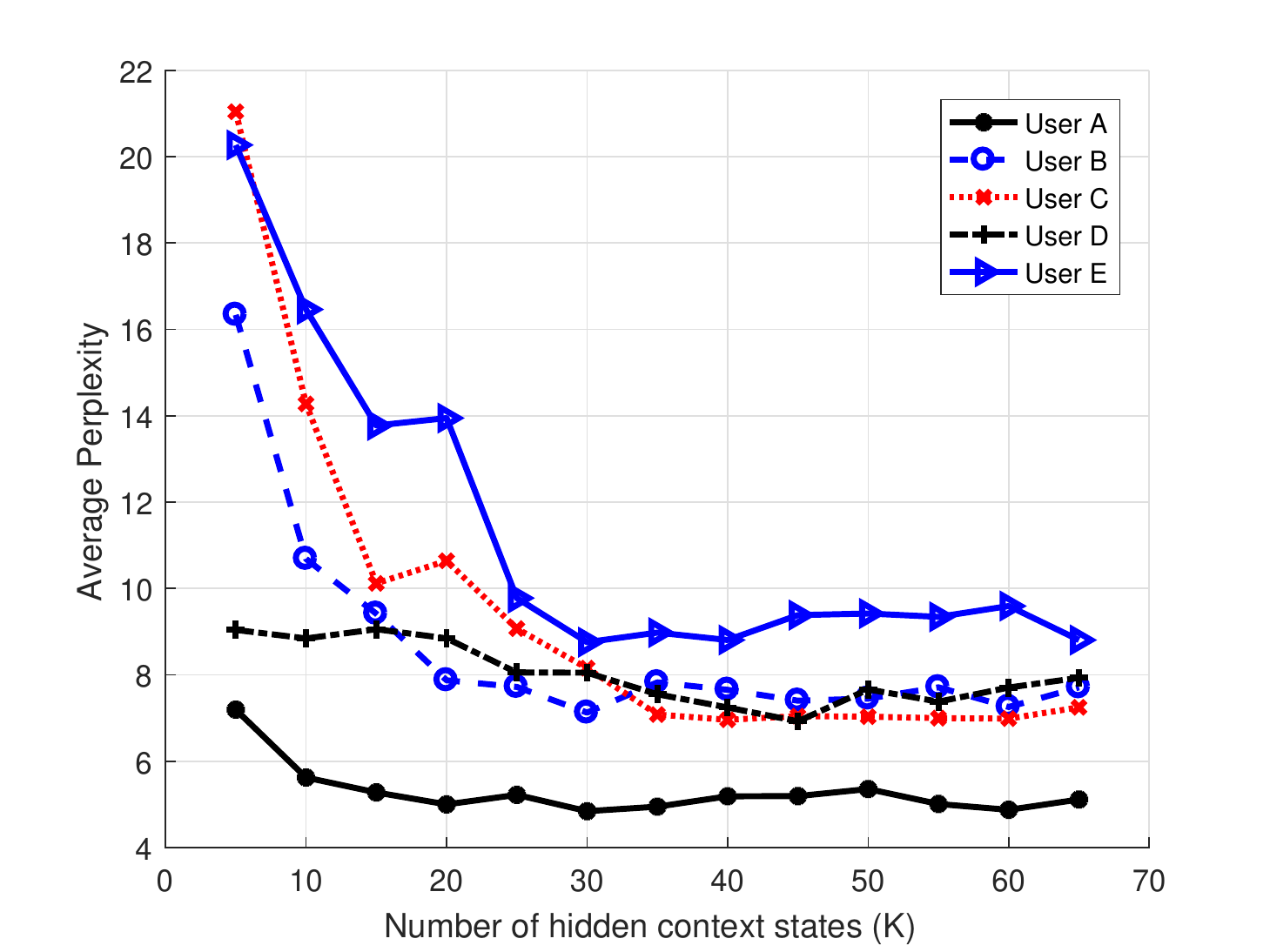}
        		\caption{}
        		\label{fig:perp_1user}
        	\end{subfigure}
~
        \begin{subfigure}[b]{0.32\textwidth}  
            \centering 
            \includegraphics[width=1.15\textwidth,height=1.8in]{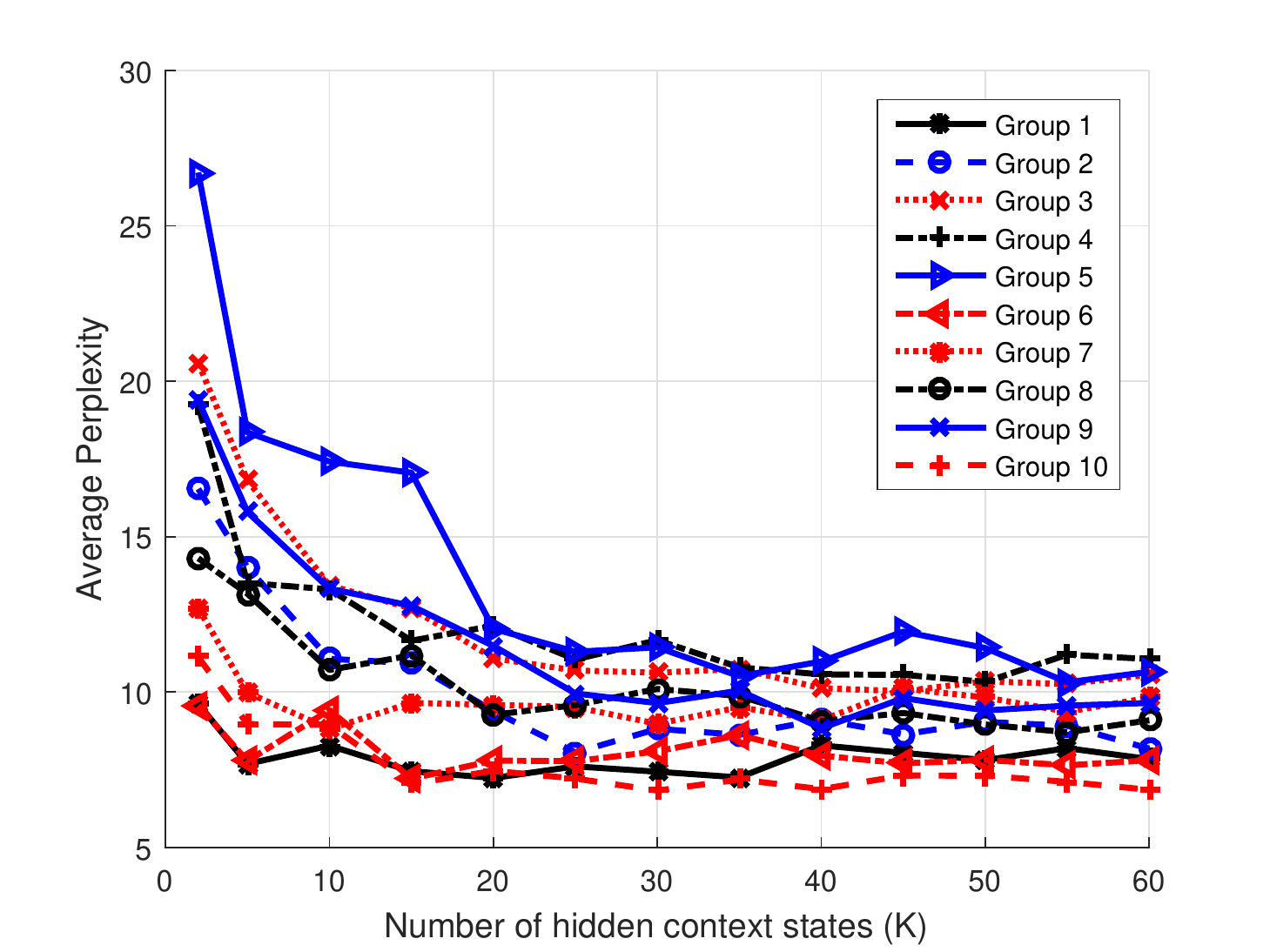}
            \caption{}
            \label{fig:perp_2user}
        \end{subfigure}
~
        \begin{subfigure}[b]{0.32\textwidth}   
            \centering 
            \includegraphics[width=1.15\textwidth,height=1.8in]{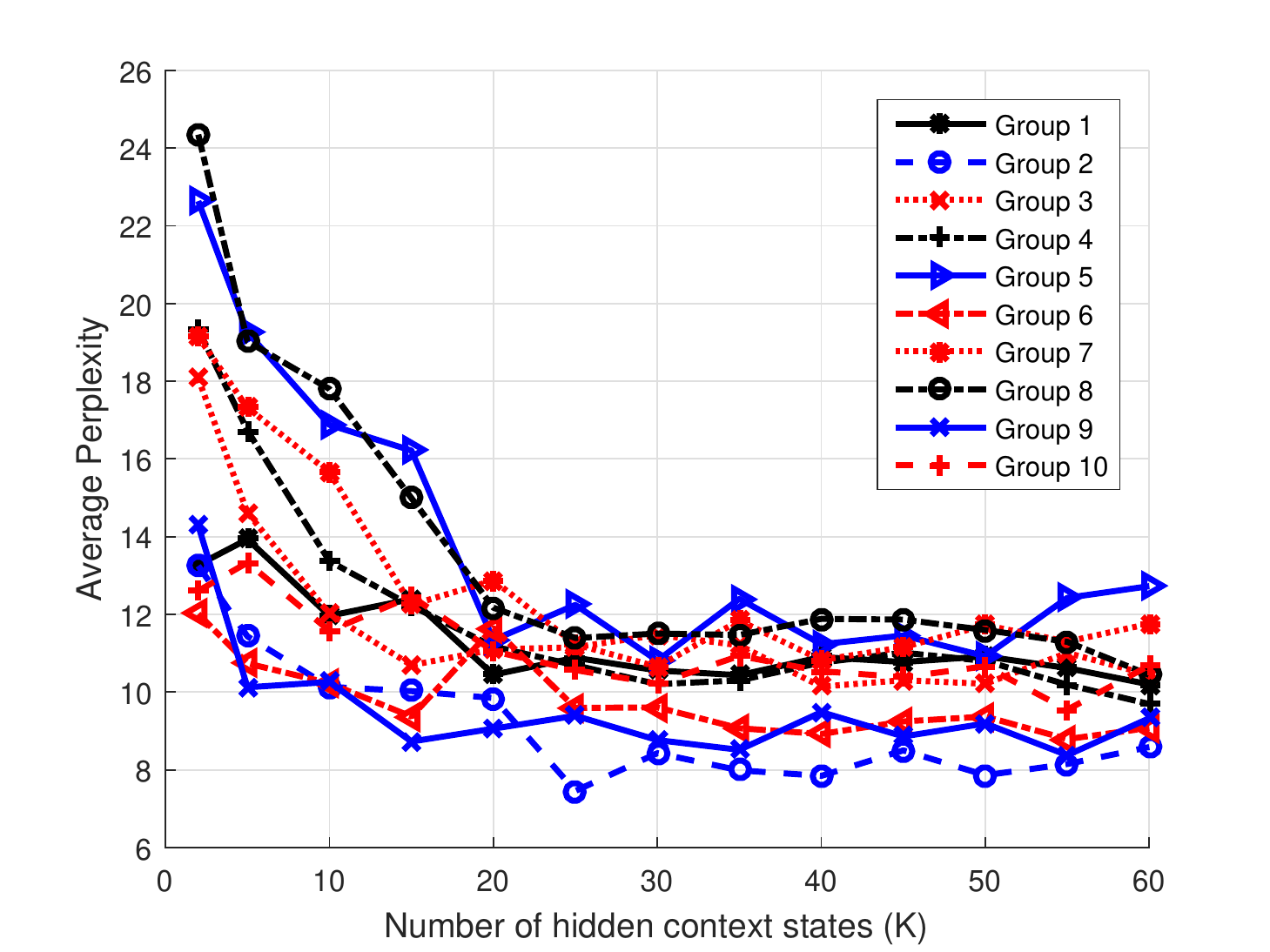}
            \caption{}
            \label{fig:perp_3user}
        \end{subfigure}
        \vspace{-0.1in}
        \caption{\label{fig:3dmap1} Perplexity vs. $K$  (a) 1 user case, (b) 2 user case, (c) 3 user case. These figures help identify the best 2 or 3 user groups (most related users).}
        \vspace{-0.15in}
\end{figure*}

\begin{figure*}[t!]
        \centering 
        \hspace{-0.3in}
           \begin{subfigure}[b]{0.32\textwidth}
        		\centering
        		\includegraphics[width=1.1\textwidth,height=1.8in]{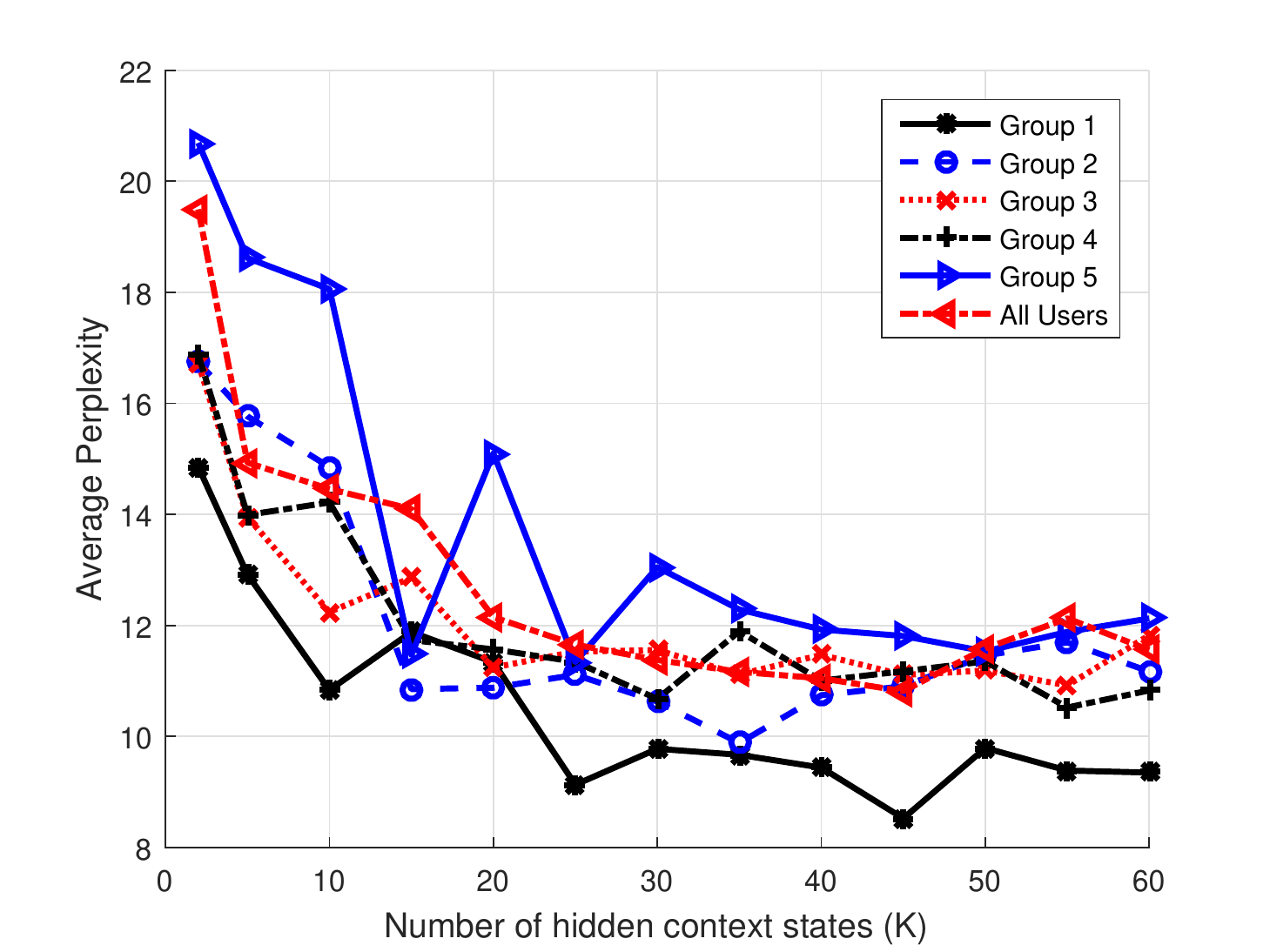}
        		\caption{}
        		\label{fig:perp_4_5user}
        	\end{subfigure}
~
        \begin{subfigure}[b]{0.32\textwidth}  
            \centering 
            \includegraphics[width=1\textwidth,height=1.8in]{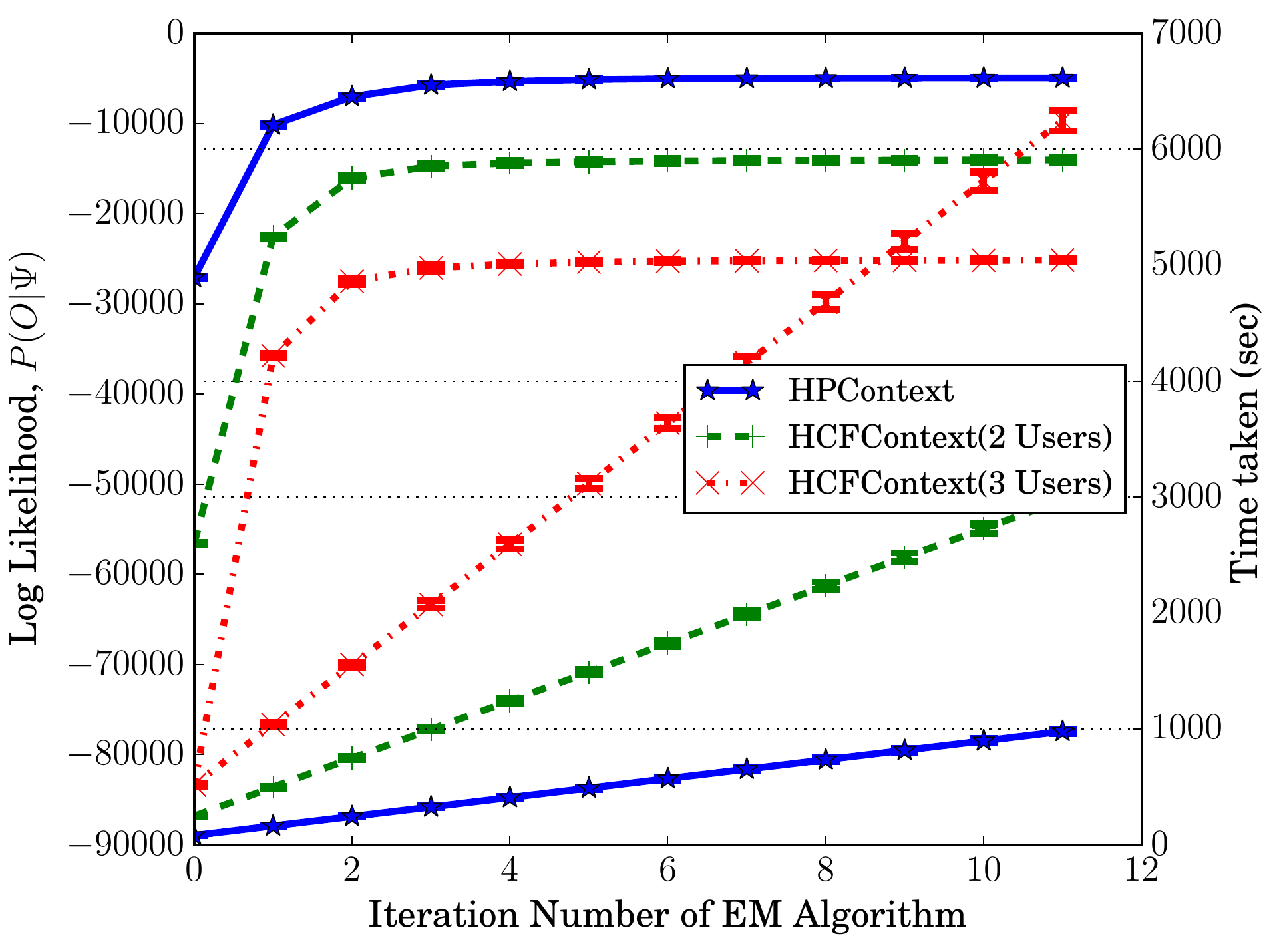}
            \caption{}
            \label{fig:lls_time_itr_num}
        \end{subfigure}
~
        \begin{subfigure}[b]{0.32\textwidth}   
            \centering 
            \includegraphics[width=1.1\textwidth,height=1.8in]{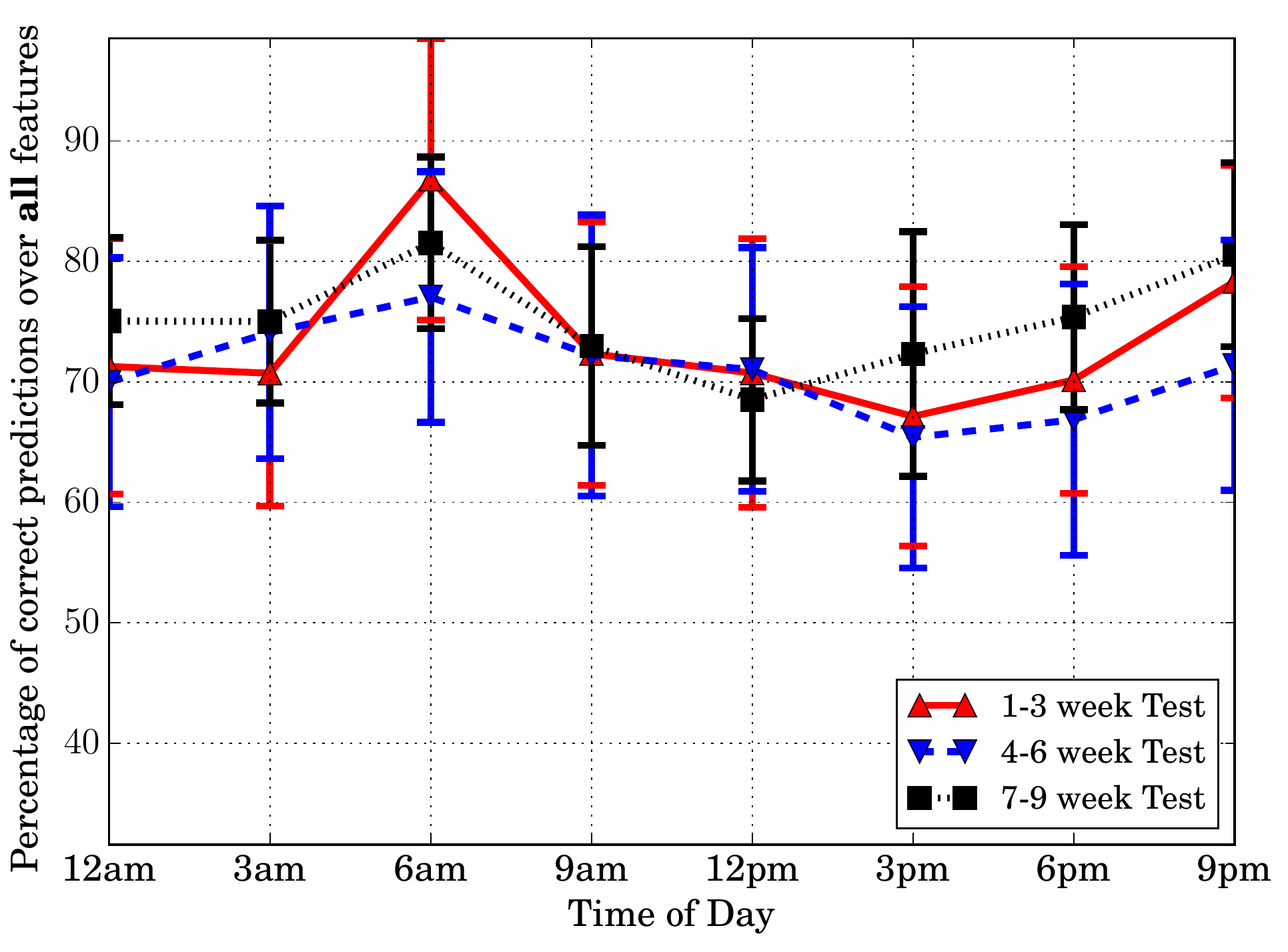}
            \caption{}
            \label{fig:cross_validation}
        \end{subfigure}
        \vspace{-0.1in}
        \caption{\label{fig:3dmap2} (a) Perplexity vs. $K$ (4 and all user case). (b) Log likelihood of the training data and time taken vs. number of iterations of EM algorithm for different models. (c) Comparison of performance (\pc + \cfc (2) + \cfc (3)) for different choices of test data - first, mid, and last 3 weeks.}
        \vspace{-0.2in}
\end{figure*}

\section{Experimental Evaluation}\label{sec:eval}
We describe the experimental setup and results, and evaluate the performance of the privacy-preserving algorithms.%

\begin{comment}
\begin{table}[]
\small
\centering
\caption{LifeMap dataset}
\label{table:lifemap_users}
\begin{tabular}{|l|l|l|l|}
\hline
\textbf{User name} & \textbf{Sex} & \textbf{Age} & \textbf{Number of places} \\ \hline
\hline
GS2       & M   & 20s & 163              \\ \hline
GS3       & M   & 20s & 297              \\ \hline
GS4       & F   & 20s & 209              \\ \hline
GS7       & M   & 20s & 289              \\ \hline
GS12      & M   & 20s & 376              \\ \hline
\end{tabular}
\end{table}
\end{comment}

\noindent \textbf{Dataset and Experiment Description:}
To validate our models, we have used the LifeMap dataset~\cite{Chon2011}, which is freely available online. This dataset consists of fine-grained mobility data such as WiFi fingerprints (MAC address and signal strengths of surrounding Wi-Fi APs), user-defined types of places (workplace, cafeteria, etc.), cell tower ID, etc. 
These details of 10 users are logged every 2 to 5 minutes for about two months (which is the overlap time among all users) in Seoul, Korea. The users are graduate students of the same lab in the university and as such the data suits our application.
We have chosen data corresponding to five users for a period of nine weeks for our experiments. 
Username, gender and the number of places visited in total for these five users are as follows---\textit{GS2 (A)}, M, 163; \textit{GS3 (B)}, M, 297; \textit{GS4 (C)}, F, 209; \textit{GS7 (D)}, M, 289; \textit{GS12 (E)}, M, 376. The average number of places visited per user is about $270$. However, the number of frequently visited places for each user ranges from 8 to 35 (median 20).
We also down sampled the data to 1 hour period to ease the computations i.e., the time instants $t$ and $t+1$ are separated by 1 hour. This also means the trained models will be able to capture mostly only those contexts with stay duration of the order of an hour or more. Six weeks of data is used for training and the rest three week data is used for testing. 
The list of features we have used from this dataset is shown in Table~\ref{table:lifemap_features}. Here $V_F$ corresponds to the total number of values taken by that feature. In case of `Holiday', Saturday, Sunday and any public holidays are considered as holidays. In cases where a certain feature's value is missing, we model it as if the feature is not available at that time using $\theta_{k,f}$. We have empirically set $\eta = \{1/K,..1/K\}, \omega_k = \{50/K,...50/K\}, \delta = \{1, 10\}, \lambda = \{0.01,...0.01\}$ in our experiments~\cite{Heinrich2008} ($K$ is the number of hidden context states). 
All our results (implemented in Python) are generated on an Intel 4-core i7-2600 CPU @ 3.40GHz, with 8 GB of RAM.
\textit{We did not consider a larger dataset as our main idea is to apply collaborative filtering across the user's \textit{closely} related users such as labmates, roommates, etc. Also, we could not find larger datasets with similar features.
}

\noindent \textbf{Perplexity vs. $K$ and Optimal Group Selection:} 
Figure~\ref{fig:perp_1user} shows the averaged perplexity values for different number of hidden states, $K$, for one user case. 
Since this is a one user case it corresponds to \pc. The perplexity is calculated by predicting the observations at the next 12 time instants in the test data given the preceding 12 observations on the same day. This has been done over all the days in the three week test data and the results are averaged. 
From Fig.~\ref{fig:perp_1user}, we can observe that perplexity reduces as $K$ increases with a pattern of diminishing returns. 
We can observe that user $A$ has the best perplexity and $K = 10$ provides a good balance between perplexity and complexity introduced due to higher $K$ values.
Figure~\ref{fig:perp_2user} shows similar result for 2 users. Since we have 2 users, we have a total of 10-user groups (5C2 combinations). We can observe that Group~10 consisting of users $A,C$ has the best perplexity and $K=15$ provides a good tradeoff. This means that the two users in Group~10 have more similar patterns than the users in other groups. Figure~\ref{fig:perp_3user} shows similar result for 3 users. We can observe that Group~2 (users $A,C,D$) has the best perplexity with $K=25$ providing good tradeoff. Similarly, for 4 user case, we can observe from Fig.~\ref{fig:perp_4_5user}, that $Group~1$ (users $A,B,C,D$) has best perplexity values and $K = 25$ provides a good balance. For all user case in the same figure, we can see that $K = 25$ provides a good tradeoff. \textit{Hence these results can be used to select most related users in a group of 2/3/4, etc. Even though it takes time to find the optimal group via this approach, we feel it is acceptable as it is done only once offline.} For the results below, whenever a user or a group of users is mentioned, we considered the above \textit{best} groups and optimal $K$ values. To illustrate the benefits of collaborative filtering contexts, we considered only 2 and 3 user groups as benefits diminish with increase in group size.

\noindent \textbf{Log Likelihood Convergence, Training Times:}
Figure~\ref{fig:lls_time_itr_num} shows the log likelihood (LL) of the training data, $P(\bm{O \mid \Psi})$, versus 
the iteration number in the EM algorithm (which is used to estimate parameters of \cfc) for different models.
We have considered six weeks of training data so, $T = 6\times 7 \times 24$. We can see that the LLs have converged and the model significantly improves the initial LL values (upto 30\%). We can notice that LLs have approximately converged after $n_{con}=3$ iterations. Even then, the converged LLs have lower values due to large number of values for certain features (such as WiFi APs which has about 440 unique values) which makes the value probabilities, $\phi_{k,f,v}$, very small.
In fact, we encountered underflow problem due to multiplication of several small probabilities and then using such a value in the denominator resulting in $nan$ values. In order to solve this problem, we have used scaling approach~\cite{Blunsom2004} for $\alpha, \beta$ values. Figure~\ref{fig:lls_time_itr_num} also shows the time taken in seconds vs. iterations of EM algorithm. The corresponding time for $n_{con}=3$ is about $200,~800,~1500$ seconds respectively for the three models. \textit{These durations are reasonable considering that the training is run offline and very less often.} Both these results are averaged over 4 runs and we can notice that the $95\%$ confidence intervals are too minute to be noticed.

\begin{table*}[t]
\footnotesize
\centering
\caption{Use Case Illustrating the Predictions (With Corresponding Max. Probabilities) at 1 Pm on One Tuesday in the Test Period.}
\label{table:usecase}
\vspace{-0.1in}
\setlength\tabcolsep{2pt} %
\begin{tabular}{|p{1.4cm}|p{1.9cm}|p{2.75cm}|p{2.75cm}|p{2.75cm}|p{2.75cm}|p{2.75cm}|}
\hline
\textit{\textbf{Features}} & \textit{\textbf{Ground Truth(A)}} & \textit{\textbf{HPC(A)}} & \textit{\textbf{HCFC(A,C)}} & \textit{\textbf{HCFC(A,C,D)}} & \textit{\textbf{HPC(A) + HCFC(A,C)}} & \textit{\textbf{HPC(A) + HCFC(A,C) + HCFC(A,C,D)}} \\ \hline \hline
Wi-Fi AP & 00:bl:f2:9b:05:76 & \textbf{00:og:1f:2e:4n:6c(0.32)} & 00:bl:f2:9b:05:76(0.43) & 00:bl:f2:9b:05:76(0.41) & 00:bl:f2:9b:05:76(0.27) & 00:bl:f2:9b:05:76(0.32) \\ \hline
Place Name & F007 & F007 (0.84) & \textbf{{B003 (0.48)}} & \textbf{{J023 (0.31)}} & F007 (0.57) & F007 (0.41) \\ \hline
Cell ID & 42534164 & 42534164 (0.62) & 42534164 (0.44) & \textbf{{48759836 (0.3)}} & 42534164 (0.53) & 42534164 (0.39) \\ \hline
LAC & 8513 & \textbf{{9353 (0.32)}} & 8513 (0.47) & 8513 (0.57) & \textbf{{9353 (0.36)}} & 8513 (0.41) \\ \hline
Battery Level & High & High (0.89) & \textbf{{Medium (0.41)}} & \textbf{{Medium (0.32)}} & High (0.57) & High (0.42) \\ \hline
Battery \mbox{Status} & Discharging & Discharging (0.95) & Discharging (0.93) & Discharging (0.81) & Discharging (0.94) & Discharging (0.89) \\ \hline
Day Period & Afternoon & Afternoon (0.72) & Afternoon (0.82) & Afternoon (0.73) & Afternoon (0.77) & Afternoon (0.75) \\ \hline
Day Name & Tuesday & Tuesday (0.59) & Tuesday (0.68) & Tuesday (0.73) & Tuesday (0.63) & Tuesday (0.67) \\ \hline
Holiday & No & \textbf{{Yes (0.6)}} & No (0.7) & No (0.75) & No (0.55) & No (0.62) \\ \hline
\end{tabular}
\vspace{-0.1in}
\end{table*}

\begin{figure*}[t!]
        \centering   
           \begin{subfigure}[b]{0.32\textwidth}
        		\centering
        		\includegraphics[width=1\textwidth]{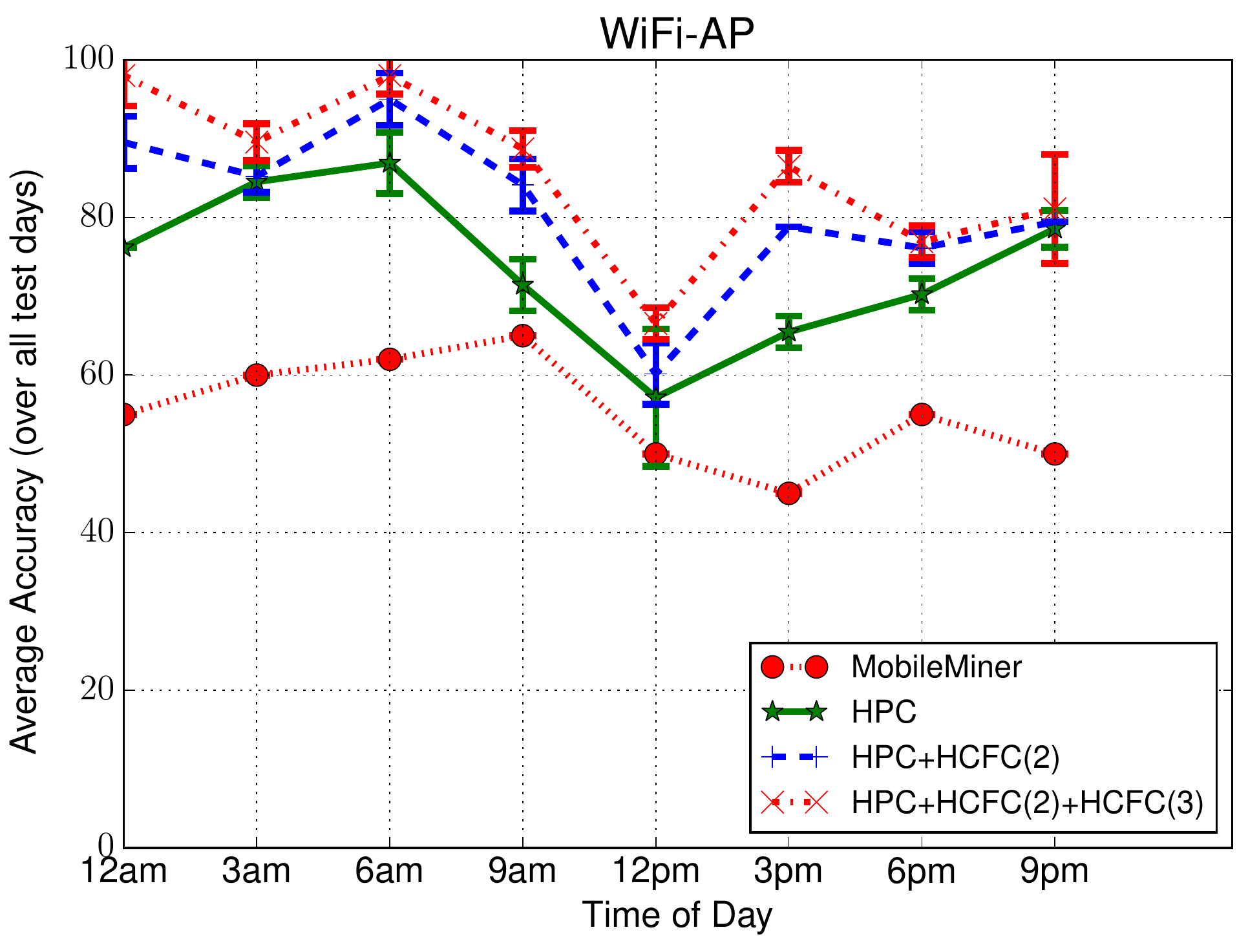}
        		\caption{}
        		\label{fig:acc-tod-wifi}
        	\end{subfigure}
~
        \begin{subfigure}[b]{0.32\textwidth}  
            \centering 
            \includegraphics[width=1\textwidth]{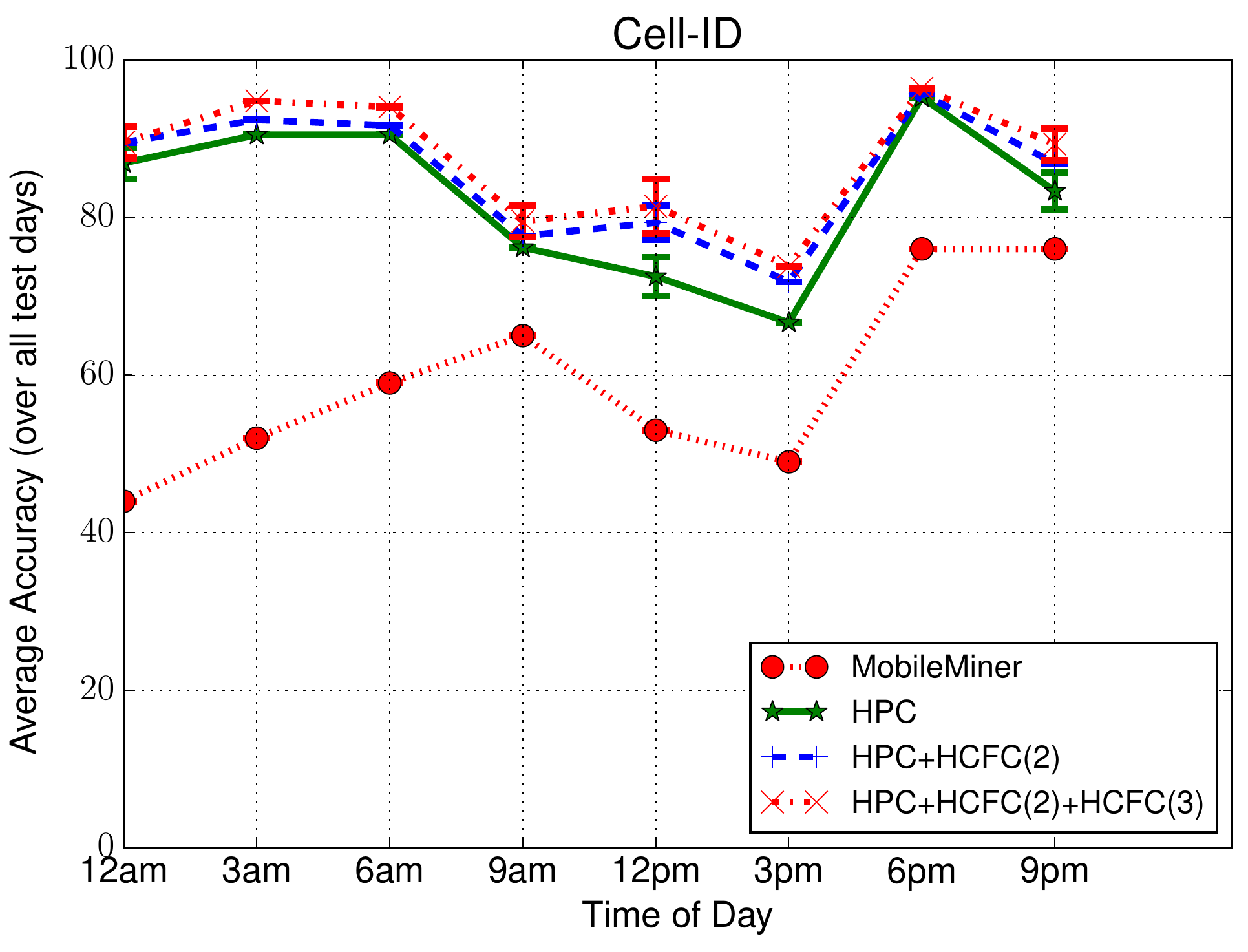}
            \caption{}
            \label{fig:acc-tod-cellid}
        \end{subfigure}
~
        \begin{subfigure}[b]{0.32\textwidth}   
            \centering 
            \includegraphics[width=\textwidth]{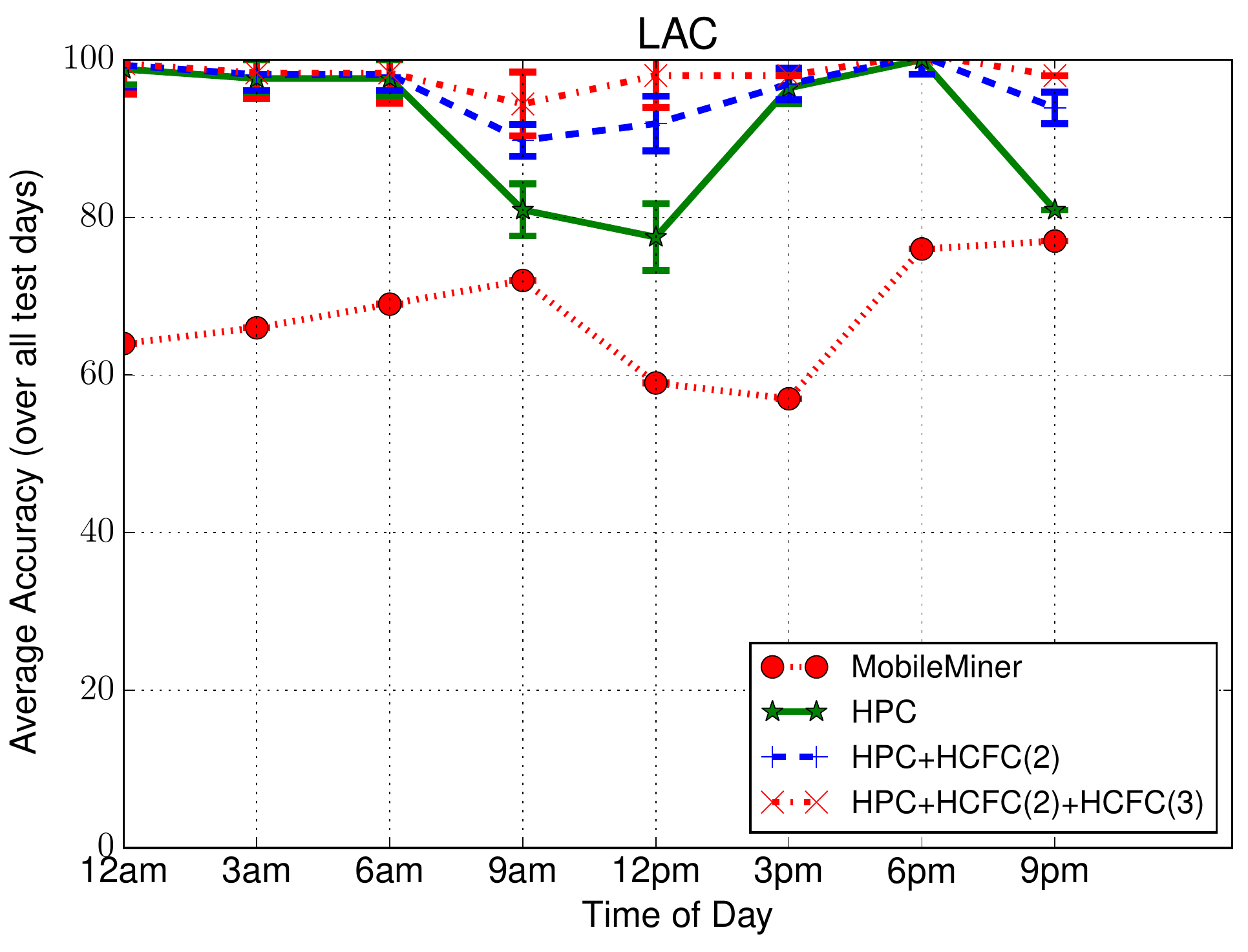}
            \caption{}
            \label{fig:acc-tod-lac}
        \end{subfigure}

           \begin{subfigure}[b]{0.32\textwidth}
        		\centering
        		\includegraphics[width=1\textwidth]{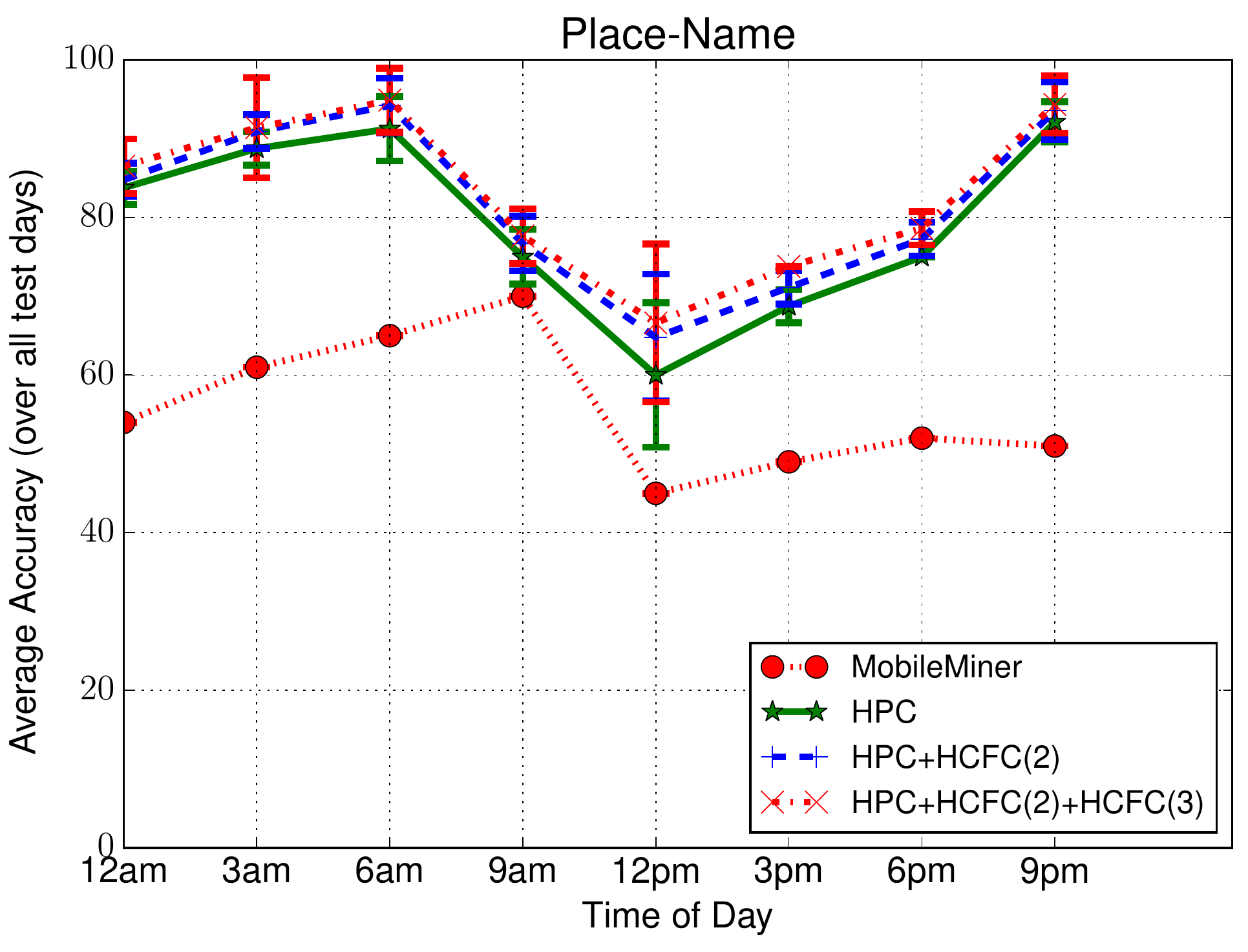}
        		\caption{}
        		\label{fig:acc-tod-placename}
        	\end{subfigure}
~
        \begin{subfigure}[b]{0.32\textwidth}  
            \centering 
            \includegraphics[width=1\textwidth]{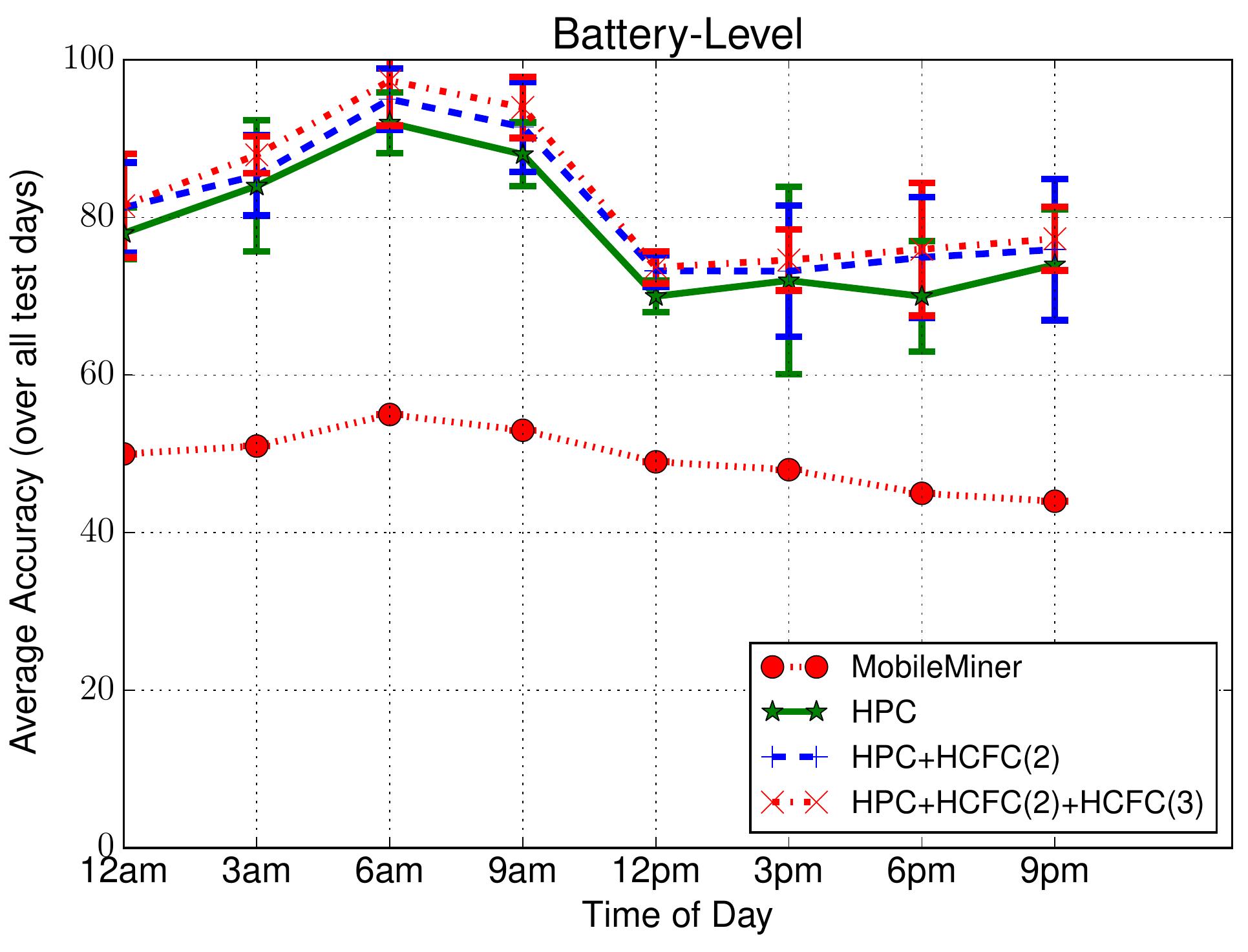}
            \caption{}
            \label{fig:acc-tod-battlevel}
        \end{subfigure}
~
        \begin{subfigure}[b]{0.32\textwidth}   
            \centering 
            \includegraphics[width=\textwidth]{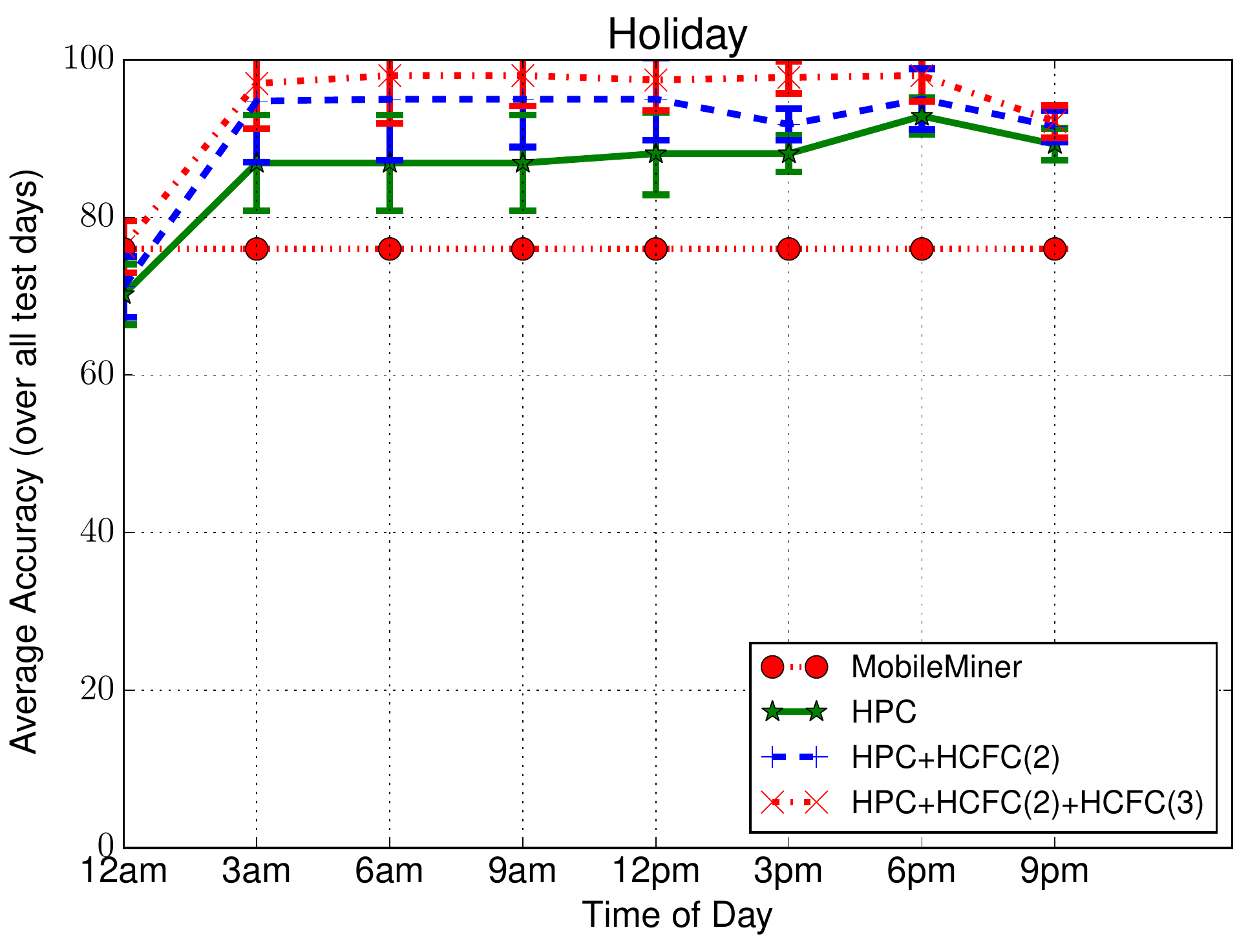}
            \caption{}
            \label{fig:acc-tod-holiday}
        \end{subfigure}
		\vspace{-0.25in}
        \caption{\label{fig:accuracyTimeOfDay} Average accuracy of the proposed models---\pc, \pc + \cfc(2), \pc + \cfc(2) + \cfc(3) in predicting the respective features at different times in a day (averaged over all 21 test days/4 runs). We can notice a slight drop in accuracy during mid-day (good at other times) owing to a degree of randomness in users motion patterns at those times (still $\approx 75\%$ sufficient for context validation/enhancement).
        }%
        \vspace{-0.2in}
\end{figure*}

\noindent \textbf{Prediction Performance---A Use Case:}
We relate to the ``lunch'' use case mentioned in Sect.~\ref{sec:intro}. We considered user $A$ for this purpose and his most related 2 and 3 user groups as found from above---$(A,C)$ and $(A,C,D)$---and the corresponding models, \pc, \cfc(2), and \cfc(3), respectively (please note this notation to be used in rest of the paper). We first illustrate the prediction results using a known use case as follows. We manually observed from the data that users $A,C,D$ usually go to lunch together on weekdays around $12-2$ pm. We wanted to check whether our models are able to capture this group behavior. Hence we predicted the contextual feature-values of user $A$ at 1 pm on one randomly selected weekday (Tuesday) in the test period given the observations of previous one-day duration using the above models. Table~\ref{table:usecase} shows the results of different models along with ground truth (column 2). All entries correspond to maximum probability values with those probabilities shown in brackets. Incorrect predictions are depicted in bold. We can notice that \pc (col. 3) does a good job in predicting the personalized features such as \textit{Place Name}, \textit{Battery Level} but makes 3 incorrect predictions for more general features such as \textit{LAC}, \textit{Holiday}, etc. Note that each feature is predicted independently, e.g., predicting \textit{Place Name} correctly does not necessarily mean \textit{Wi-Fi AP} prediction is correct. Interestingly, \cfc (cols. 4,5) makes correct predictions for those general features but fares badly for personalized features. Hence we combined the two models to obtain better predictions (as can be seen in last two columns) as follows. In case of \pc + \cfc(2), we first average the probabilities of values predicted by both models and then take the feature value with the highest average probability; similarly for \textit{HPC} + \textit{HCFC}(2) + \textit{HCFC}(3) (\textit{HPC} refers to \pc and \textit{HCFC} to \cfc for simplicity).

\begin{figure*}[t!]
        \centering   
           \begin{subfigure}[b]{0.32\textwidth}
        		\centering
        		\includegraphics[width=1\textwidth]{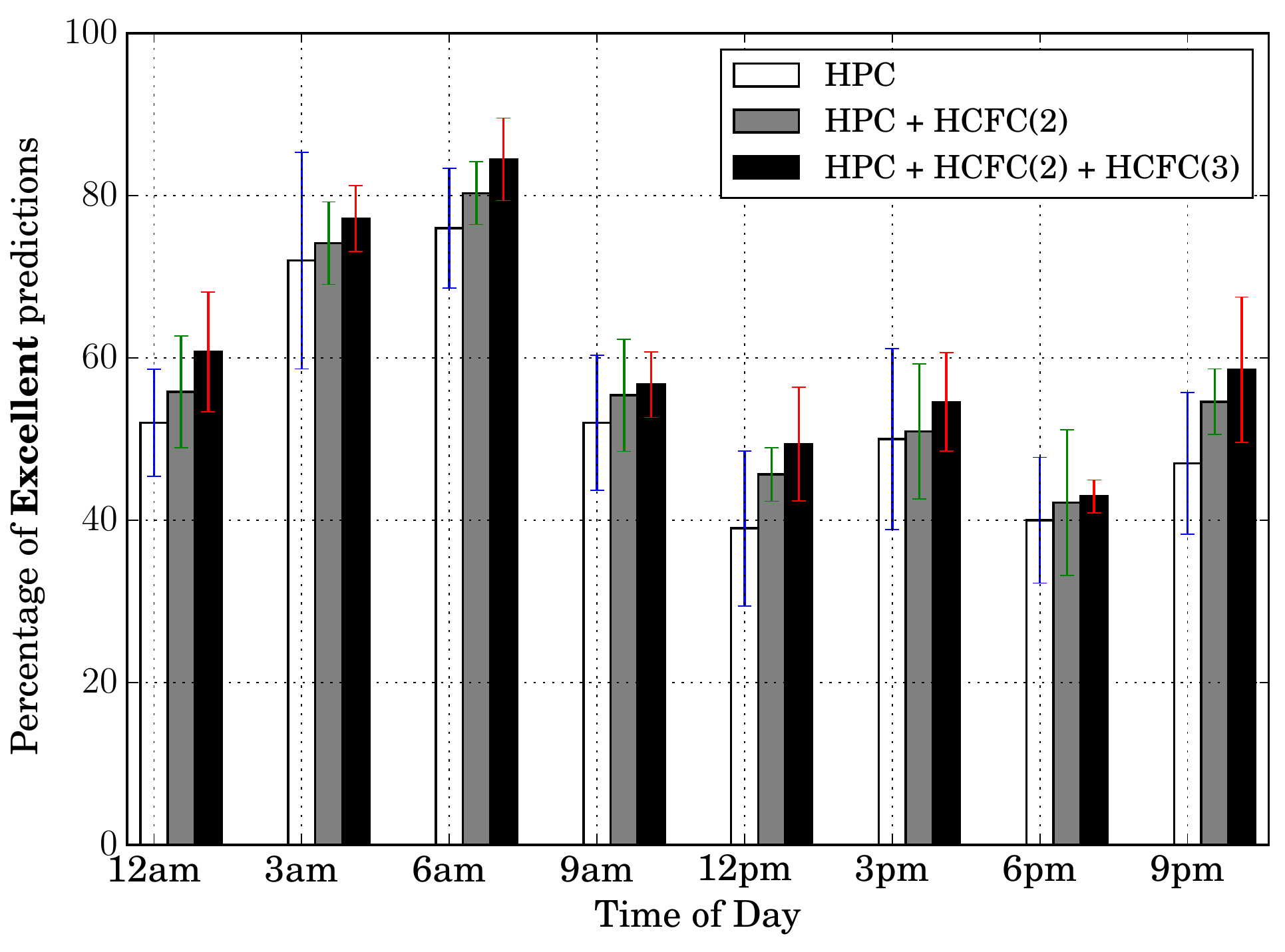}
        		\caption{}
        		\label{1}
        	\end{subfigure}
~
        \begin{subfigure}[b]{0.32\textwidth}  
            \centering 
            \includegraphics[width=1\textwidth]{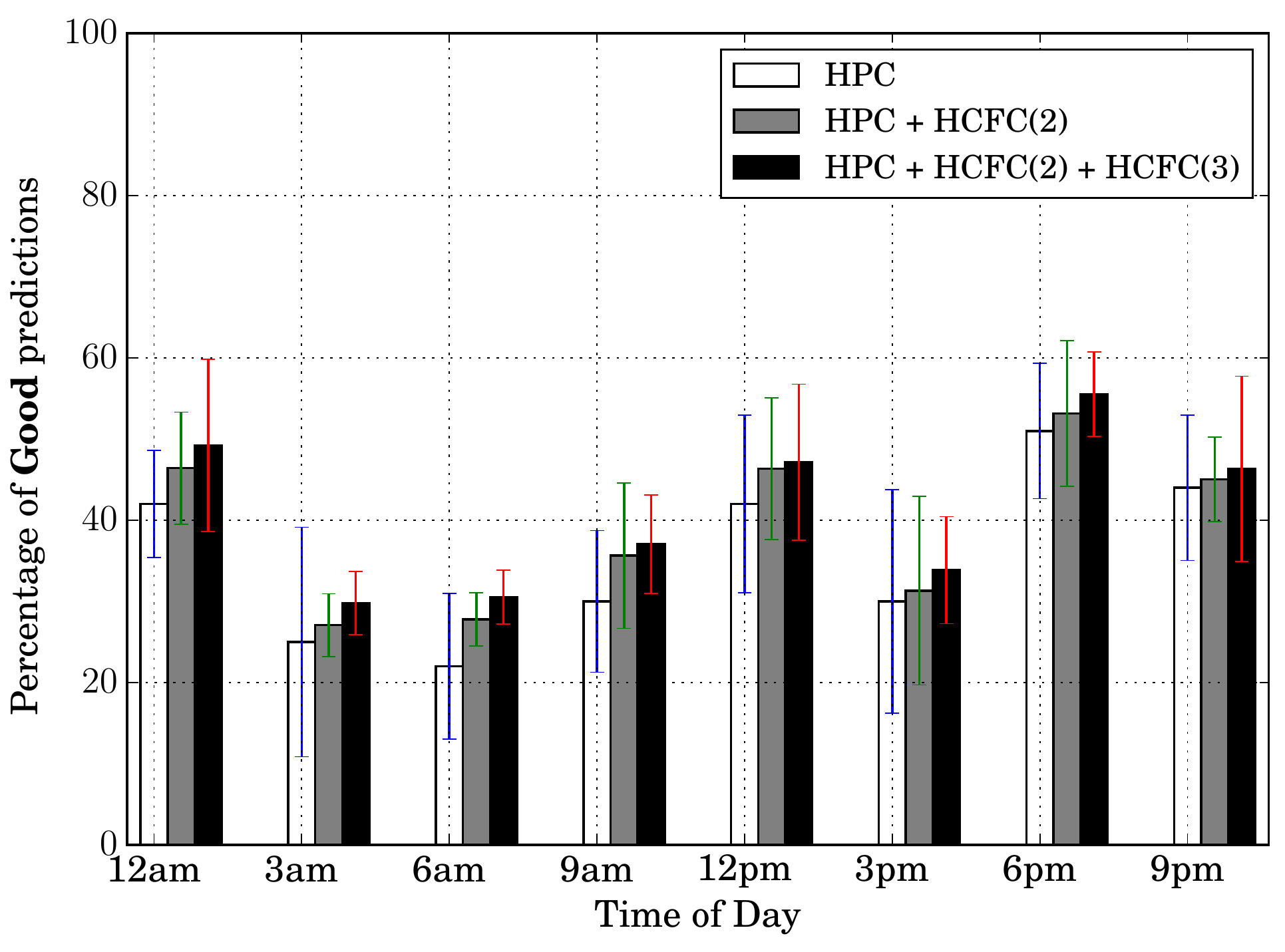}
            \caption{}
            \label{2}
        \end{subfigure}
~
        \begin{subfigure}[b]{0.32\textwidth}   
            \centering 
            \includegraphics[width=1\textwidth]{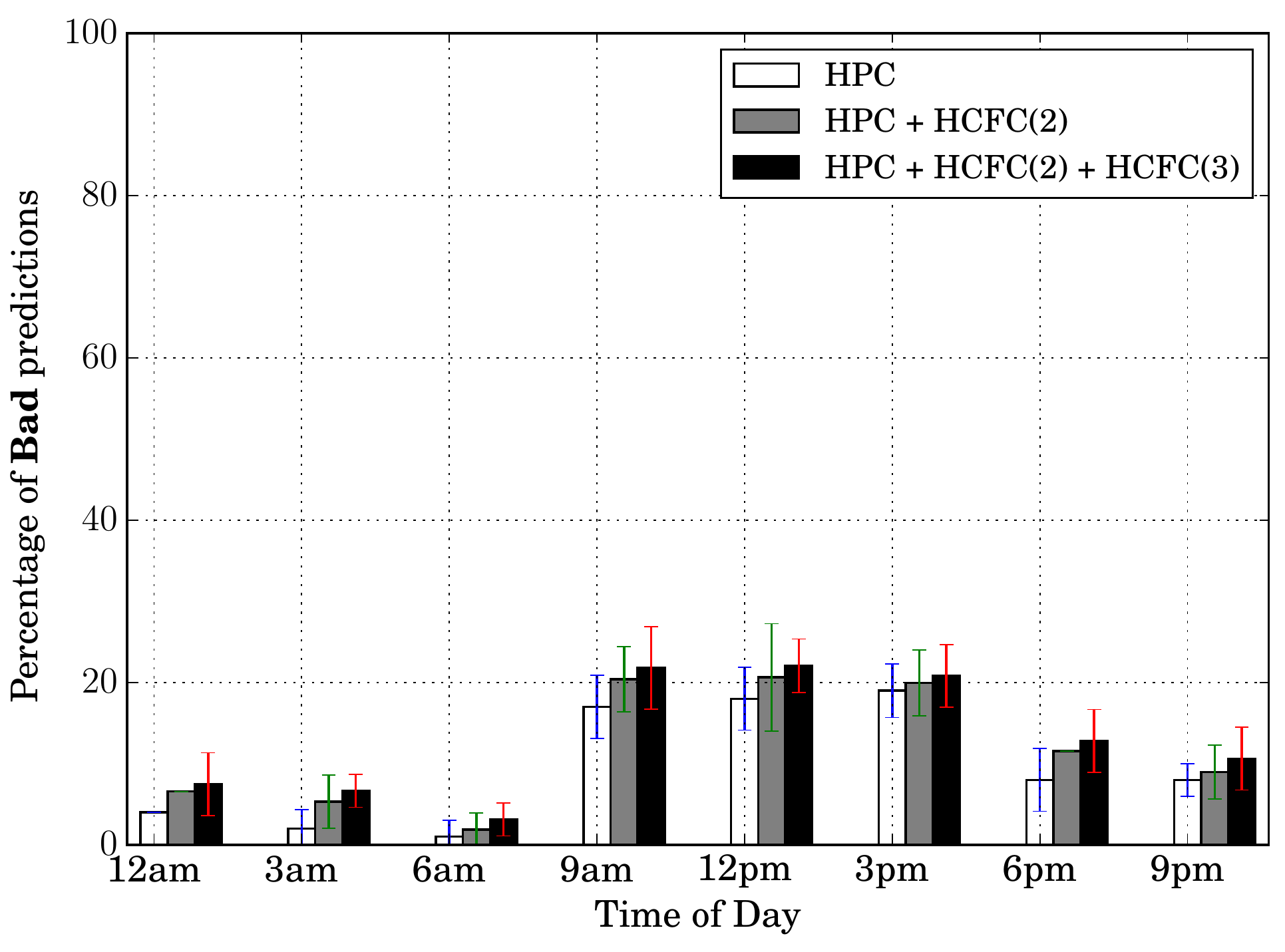}
            \caption{}
            \label{3}
        \end{subfigure}
~
        \begin{subfigure}[b]{0.32\textwidth}
        		\centering
        		\includegraphics[width=1\textwidth]{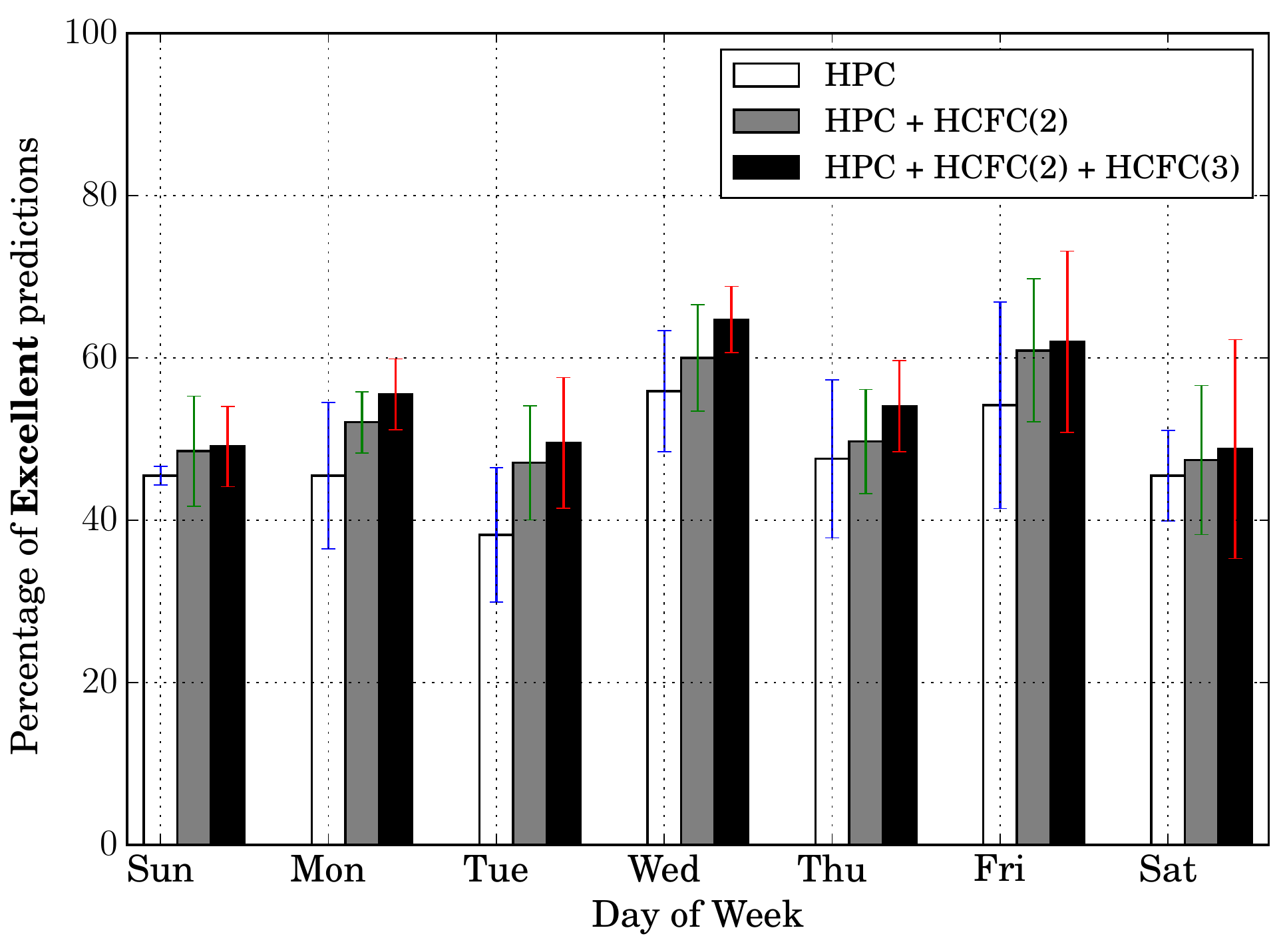}
        		\caption{}
        		\label{4}
        	\end{subfigure}
~
        \begin{subfigure}[b]{0.32\textwidth}  
            \centering 
            \includegraphics[width=1\textwidth]{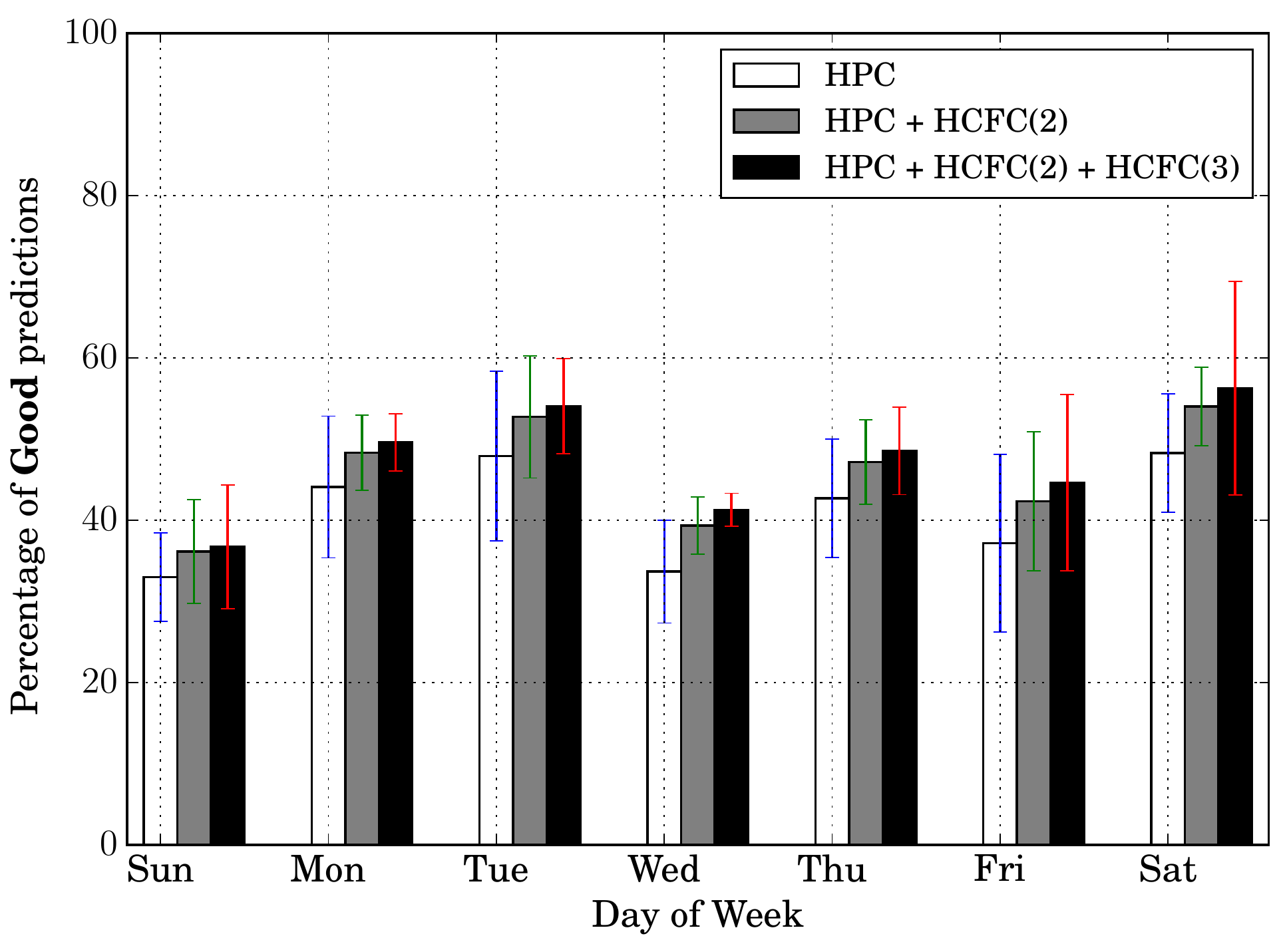}
            \caption{}
            \label{5}
        \end{subfigure}
~
        \begin{subfigure}[b]{0.32\textwidth}   
            \centering 
            \includegraphics[width=1\textwidth]{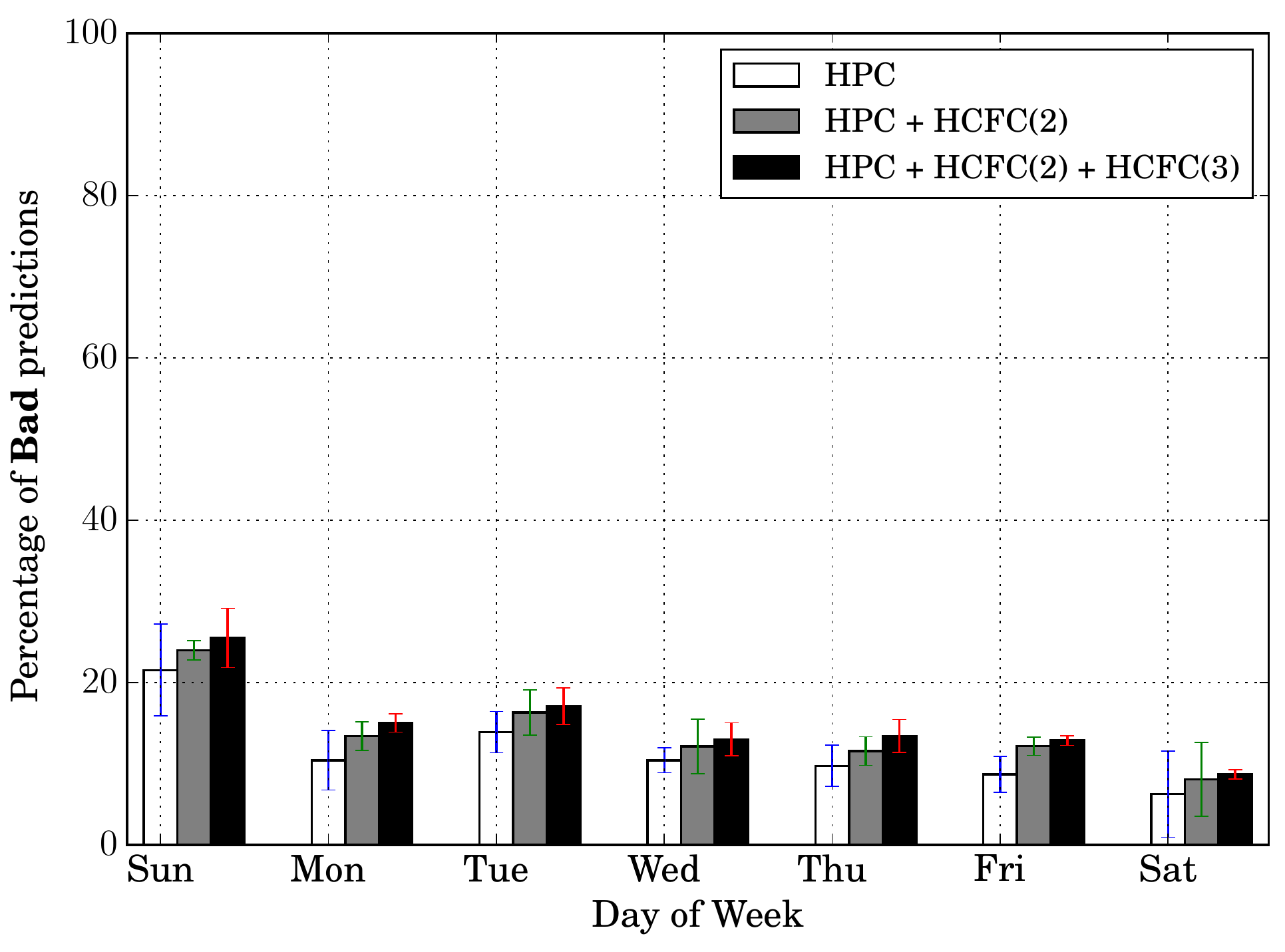}
            \caption{}
            \label{6}
        \end{subfigure}
        \vspace{-0.25in}
        \caption{\label{fig:exce_good_bad_cases} Percentage of excellent/good/bad predictions---(a)-(c) at different times of day; (d)-(f) for different days of week.}
        \vspace{-0.15in}
\end{figure*}

\noindent \textbf{Prediction Performance---Overall:} To evaluate the overall performance, we predicted all the contextual feature value pairs of user $A$ at 3 hour increments in the entire 3-week testing period (i.e., $3\times 7 \times 8$ in total) given the past observations of one-day duration. In order to compare with other closest approaches, we considered Mobile Miner~\cite{Srinivasan2014}, which is the current state-of-the-art machine learning algorithm to mine contextual co-occurrences. Each feature is considered to be independently co-occurring with the time of day and day of week (as opposed to sequentially occurring in our case), and is modeled using Multinomial Logistic Regression. Figure~\ref{fig:accuracyTimeOfDay} shows the average prediction accuracy (percentage of correct predictions) for each feature at different times in a day (averaged over all the $21$ days). Even though we are able to predict at the hourly level (due to the upsampling mentioned earlier), we show only at 3-hour intervals for clarity. These simulations are also run for 4 runs (to account for randomicities such as random initialization of the model parameters) and the confidence intervals are shown. 
Due to space limitations, we have shown results only for six features consisting of four location features---Wi-Fi AP, Place Name, Cell ID, Location Area Code~(LAC), one device feature---Battery Level and one time feature---Holiday. Others follow similar pattern. 

\noindent First of all we can notice that the proposed models perform better than Mobile Miner.
Second, in case of proposed models, we can notice that the accuracy slightly drops during mid-day compared to other times. The reasons for this drop as follows---(1) the user is more mobile during those times introducing randomness into the data making the model hard to learn; (2) the average number of places visited by each user is $270$ (as indicated above) and the $K$ value chosen (as a tradeoff between complexity and accuracy) is less than that; (3) the data is down-sampled to one hour interval, meaning, any places with stay duration less than one hour will not be captured well. However it is important to note that the accuracy in such cases is still on an average about 75\% which is reasonable to validate/complement the context from sensors. During other times of the day, we can notice that the accuracy is about $90\%$, which means the models are able to predict the values of those contextual features correctly 90\% of the time. Third, we can notice that \textit{HPC} + \textit{HCFC}(2) model improves accuracy over \textit{HPC} to a maximum of $15\%$ especially in more general features such as \textit{LAC}, \textit{Holiday}. In case of personalized features such as \textit{Place Name}, \textit{Battery Level}, the improvement is minor. We also notice that the additional improvement from \textit{HCFC}(3) is again minor, about $5\%$. In total, \textit{HCFC} contributes upto $20\%$ improvement in accuracy.
We have also plotted similar results for each day of the week but could not include them due to space limitations. 
In addition to above trends in generic/personalized features vs. models, we have noticed a drop in accuracy over weekends (to $\approx 75\%$ on average) again owing to increased mobility with less sequential behavior. The average accuracy is about $80\%$ to $90\%$ for the rest of the days (\textit{HPC} + \textit{HCFC}(2) + \textit{HCFC}(3) model). 

\noindent \textbf{Prediction Quality:} Next, we tested our models' prediction quality by testing how many of the features (all in Table~\ref{table:lifemap_features}) the models are able to predict correctly  at a given time of day. For this purpose, we created three categories---\textit{Excellent}, \textit{Good}, \textit{Bad}. If the model is able to predict at least 7 features out of 9 correctly at a given instant, we call it \textit{Excellent} prediction. Similarly we call 4/5/6 features prediction, a \textit{Good} prediction and 1/2/3 ($0$ is not included) features prediction, a \textit{Bad} prediction. For example, the prediction corresponding to $HPC(A)$ in Table~\ref{table:usecase} is considered a \textit{Good} prediction, while that belonging to $HPC(A)+HCFC(A,C)$, an \textit{Excellent} prediction.
Figure~\ref{fig:exce_good_bad_cases} shows these results (averaged) for different times of day and days of week. In both sets of figures, we can notice that the percentage of \textit{Excellent} cases is at least $50\%$. Secondly, the percentage of \textit{Excellent} cases is more than \textit{Good} cases which in turn is more than the \textit{Bad} cases (in particular the \textit{Bad} cases are very less comparatively).

\begin{figure*}[t!]
        \centering   
           \begin{subfigure}[b]{0.32\textwidth}
        		\centering
        		\includegraphics[width=1\textwidth,height=1.8in]{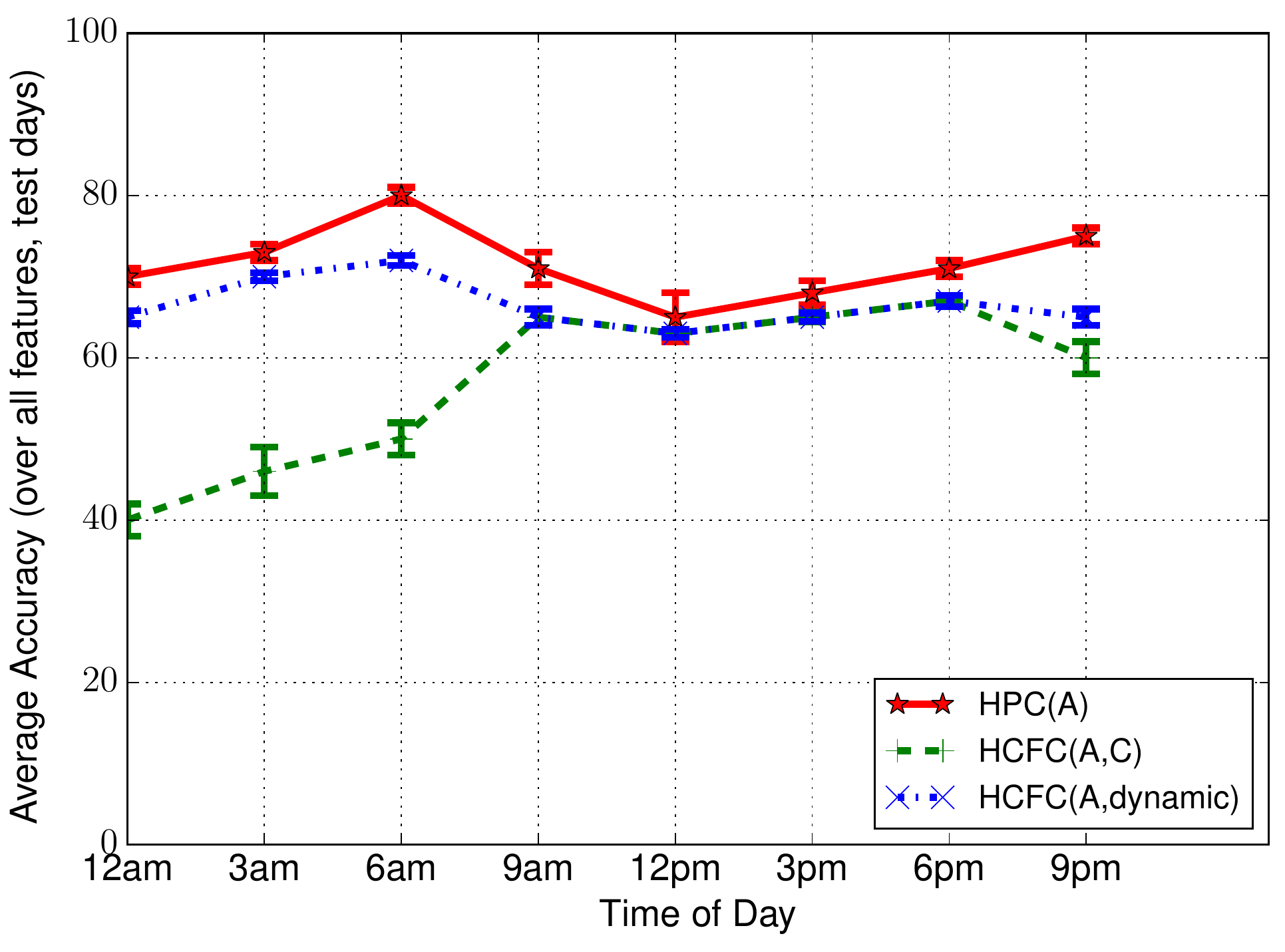}
        		\caption{}
        		\label{fig:acc_dyn_user_group}
        	\end{subfigure}
~
        \begin{subfigure}[b]{0.32\textwidth}  
            \centering 
            \includegraphics[width=1\textwidth, height=1.6in]{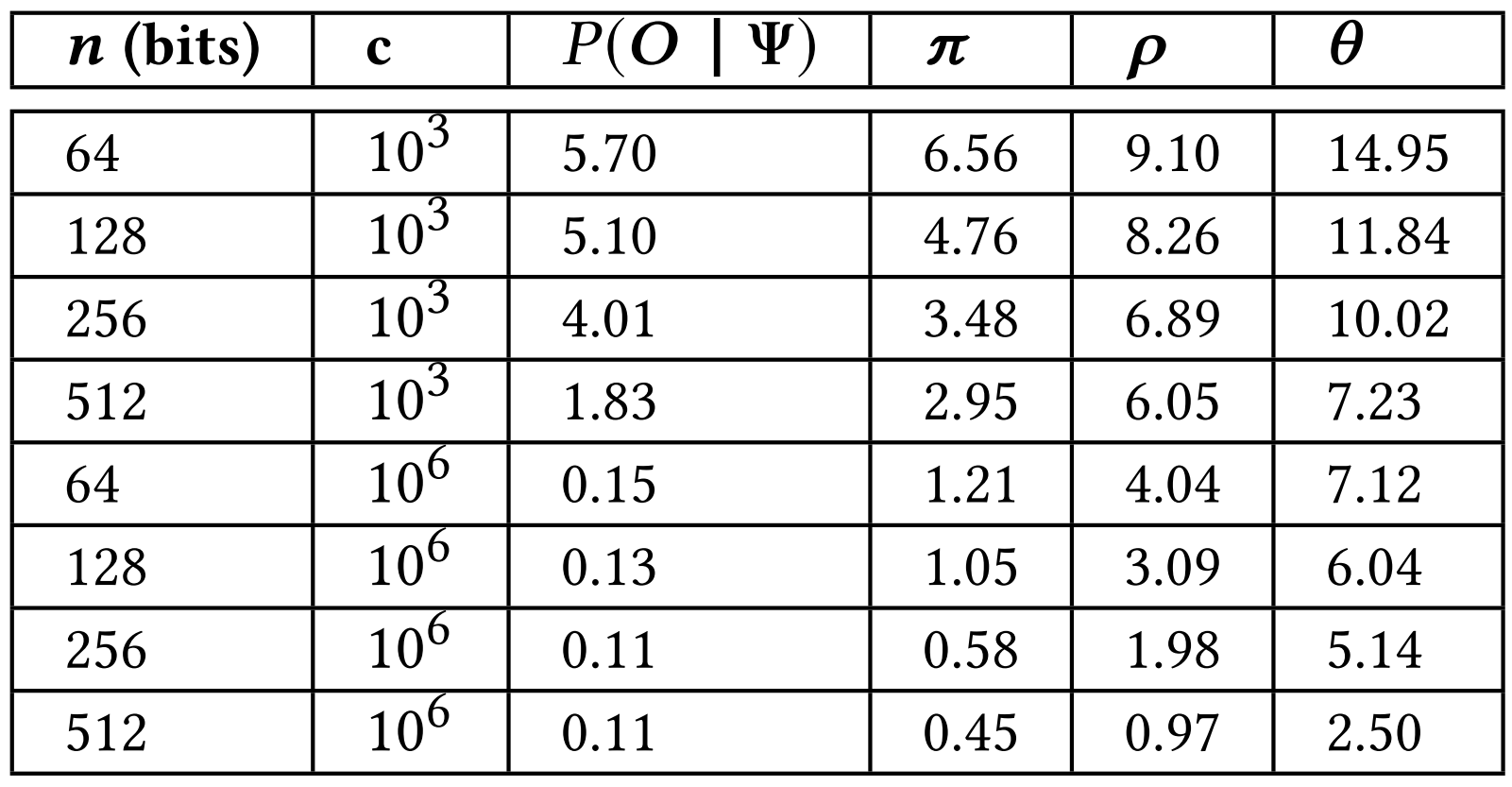}
            \vspace{0.01in}
            \caption{}
            \label{fig:pp_errors}
        \end{subfigure}
~
        \begin{subfigure}[b]{0.32\textwidth}   
            \centering 
            \includegraphics[width=1\textwidth, height=1.8in]{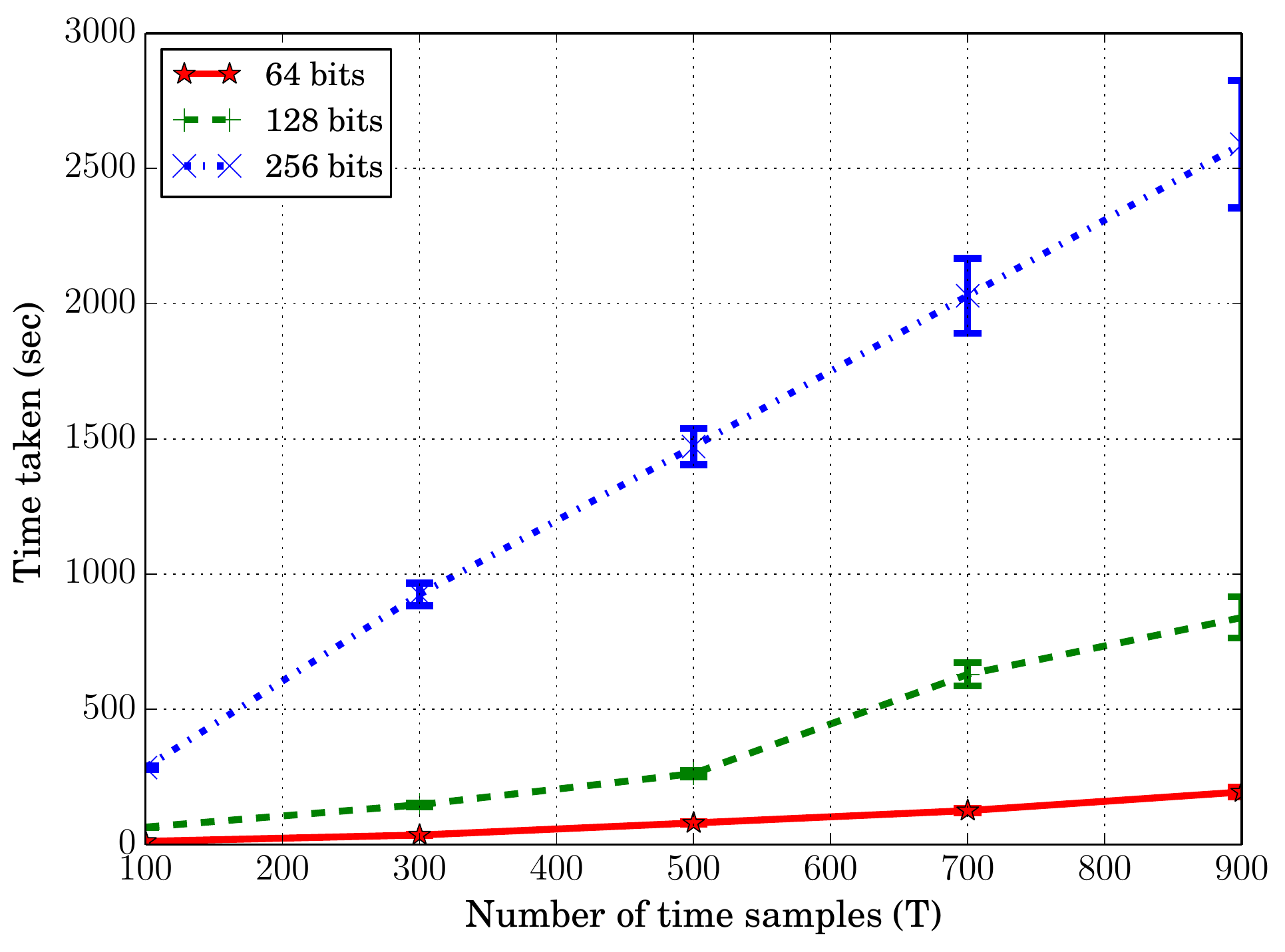}
            \caption{}
            \label{fig:pp_runtimes}
        \end{subfigure}
        \vspace{-0.25in}
        \caption{\label{fig:3dmap3} (a) Comparison of performance among \pc, \cfc (with fixed optimal user group), \cfc (with dynamic optimal user group). (b) Worst-case errors of parameters~[10 runs]~(\%). (c) Time taken to run Algorithm~\ref{alg:forward} for different number of training (time) samples vs. key length (bits).}
        \vspace{-0.2in}
\end{figure*}

\noindent \textbf{Test Data Rotation:}
The performance of the models when the test data chosen is the first (1-3), middle (4-6) and last (7-9) three weeks is shown in Fig.~\ref{fig:cross_validation}, which shows the percentage of correct predictions across all features and test days for different choices of test data (results shown only for \textit{HPC} + \textit{HCFC}(2) + \textit{HCFC}(3) for clarity). We can notice that the performance is roughly the same showing robustness of the proposed models to choice of test data and their ability to fully learn users' sequential patterns using 6 weeks train data.

\noindent \textbf{Contextual Optimal User Group Selection:} So far, for a given user, $A$, we have found the optimal user groups considering all times of the day. However, it is more beneficial to find the optimal user group based on the time of the day as is the real scenario. For example, for a given user, the closely related users during the office hours may be different from the closely related users during home hours. Taking this idea into account, we have found the most closely related user for user $A$ (i.e., optimal 2-user group) at different times of the day using the perplexity method mentioned earlier. Instead of averaging across all times of the day, we find the user group with minimum average perplexity at each time instant of the day. The results are as follows---$(12am, B)$, $(3am, B)$, $(6am, B)$, $(9am, C)$, $(12pm, C)$, $(3pm, C)$, $(6pm, C)$, $(9pm, D)$.

\noindent \textbf{Prediction Using Only \cfc}: We now predict the feature-value pairs of user $A$ at all time instants in the test period using only the collaborative filtering model, \textit{HCFC}(2), with dynamic optimal 2-user group obtained from above. Figure~\ref{fig:acc_dyn_user_group} shows the performance of that model compared against two other models---the personalized model \textit{HPC} and \textit{HCFC}(2) with fixed optimal 2-user group $(A,C)$. We can notice that dynamic \textit{HCFC} performs close to \textit{HPC} and better than fixed \textit{HCFC}. This indicates that personalized context can be obtained from collaborative filtering of contexts corresponding to user's closely related users with appropriate dynamic (i.e., time of day/activity the user is performing) selection of closely related users.

\noindent \textbf{Privacy-preserving Algorithms:}
To evaluate the performance of Algorithms~\ref{alg:forward}, \ref{alg:bw}, we tested them on a simple HMM with $M=2$ parties, $K=2$ hidden states and $|\bm{f}_{tu}=6|$ observation states per hidden state. We evaluated both the amount of error introduced (due to scaling as mentioned in Sect.~\ref{sec:prop-soln:privacy})) as well as the time taken to train the HMM and run the predictions. For the former, we calculated the error by comparing with the non-privacy preserving case. We varied the key length (bits) $n$ and the scaling factor $c$. 
The worst-case errors over 10 runs with $T=1000$ samples, as percentages, for different parameters is shown in Fig.~\ref{fig:pp_errors}. We notice that the error reduces as scaling factor increases (as expected). Similarly, as the key length increases, the error reduces and also security increases. We can notice that for large keys and reasonable scaling factors, the error due to integer approximation and consequent over- or underflow is insignificant (about 2\% in the worst case over all parameters). This shows that the effect on the prediction accuracy results above will not be drastic. However, the price is in terms of run-time. Figure.~\ref{fig:pp_runtimes} shows the average run-times of Algorithm~\ref{alg:forward} vs. the number of time samples as well as the key length (with $c=10^6$). We can see that the time taken varies linearly with the number of samples used. However, the relation seems to be approximately quadratic with key-length. This result shows the tradeoff between security and run-time. As the key-length is increased, more is the security, less is the amount of error introduced, however the run-times are more. Hence a suitable key length should be chosen that is a compromise between security/error and run-time. The run-times for Algorithm~\ref{alg:bw} are on average five times more. \textit{However these algorithms need to be run only for training the \cfc which is run very less frequently and offline.} Moreover, we will leverage recent advances in encryption and multi-party computation algorithms such as~\cite{Bitar2017} to help further reduce the runtimes of these algorithms, as part of future work.

\noindent \textbf{Real-time Inference and Cold-start:}
Once the training is complete, model parameters are known to each device. Prediction, then, just involves evaluating~\eqref{eq:featurepred}, \eqref{eq:valuepred} by plugging in the learned parameters. These are just a few arithmetic operations and do not involve any compute-intensive encryption/decryption algorithms unlike training which happens offline. \textit{Hence, our approach will not have any problems in practical implementations i.e., making predictions in real time.} Furthermore, to combat the cold-start problem akin to collaborative filtering based approaches, we suggest to---(i)~use sensor context and also obtain user validation for additional security; (ii)~use sensor context + \pc until \cfc is learnt well.

\noindent \textbf{Energy Considerations:}
Note that all of the features we worked with are passive, i.e., do not require active probing that consumes energy. One exception is the Wi-Fi, which needs to be turned ON in case it is not ON. In all other cases, we piggyback on the sensor data already available on the phone, reducing energy consumption.

\section{Conclusion and Future Work}\label{sec:conc}
We proposed and evaluated (on a real-life dataset with over $80\%$ accuracy) privacy-preserving, sequential history-based personalized and collaborative-filtering models, for current and future mobile context prediction to validate and/or enhance the sensor context. Their feasibility for practical deployment in security applications and/or mobile personal assistant technologies is shown.
\begin{comment}
We have shown that the proposed models are able to predict the correct feature-value pairs with over $80\%$ accuracy and that it is possible to predict context using collaborative filtering models alone by dynamic selection of optimal users.
We also presented privacy-preserving algorithms for multi-party parameter estimation of the proposed models
and show that they are indeed feasible candidates for practical deployment in security applications and/or mobile personal assistant technologies. 
\end{comment}
As future work, we plan to conduct a pilot study of our models; improve their training times leveraging suboptimal algorithms and enhance their prediction accuracy.

\section*{Acknowledgment}
We thank the US Department of Homeland Security Science \& Technology Directorate~(DHS S\&T) Cyber Security Division for their support under the contract No.~D15PC00159.

\balance
\newpage %

\bibliographystyle{IEEEtran}%
\bibliography{references,url-refs}

\end{document}